\documentclass{article}
\makeatletter\def\input@path{{headers/}}\makeatother
\usepackage{iclr2027_conference,times}

\pdfoutput=1

\usepackage{microtype}
\usepackage{graphicx}
\usepackage{afterpage}   
\usepackage{subcaption}  
\usepackage{booktabs}
\usepackage{url}

\usepackage{amsmath}
\usepackage{amssymb}
\usepackage{mathtools}
\usepackage{amsthm}

\usepackage{hyperref}
\definecolor{darkblue}{rgb}{0, 0, 0.5}
\hypersetup{colorlinks=true, citecolor=darkblue, linkcolor=darkblue, urlcolor=darkblue}

\usepackage[capitalize,noabbrev]{cleveref}
\crefname{assumption}{Assumption}{Assumptions}
\Crefname{assumption}{Assumption}{Assumptions}
\crefname{desiderata}{Desiderata}{Desiderata}
\Crefname{desiderata}{Desiderata}{Desiderata}

\theoremstyle{plain}

\theoremstyle{definition}

\theoremstyle{remark}

\usepackage{algorithm}
\usepackage{algorithmic}

\usepackage[disable,textsize=tiny]{todonotes}

\usepackage{xcolor}
\definecolor{WowColor}{rgb}{.75,0,.75}
\definecolor{SubtleColor}{rgb}{0,0,.50}

\newcounter{margincounter}

\usepackage[most]{tcolorbox}

\tcbset{
  basecolorbox/.style={
    enhanced,
    boxrule=0.5pt,
    arc=4pt,
    outer arc=4pt,
    boxsep=5pt,
    left=5pt,
    right=5pt,
    top=5pt,
    bottom=5pt,
    fonttitle=\bfseries,
    title={#1},
    before upper={\begin{quote}\obeylines},
    after upper={\end{quote}}
  }
}

\tcbset{
  codeboxstyle/.style={
    enhanced,
    boxrule=0.5pt,
    arc=4pt,
    outer arc=4pt,
    boxsep=5pt,
    left=5pt,
    right=5pt,
    top=5pt,
    bottom=5pt,
    fonttitle=\bfseries,
    title={#1}
  }
}

\definecolor{forecastblue}{RGB}{115,137,170}
\definecolor{contrastterracotta}{RGB}{181,101,74}
\definecolor{agreementmustard}{RGB}{193,145,68}
\definecolor{cruxgreen}{RGB}{105,165,88}

\newtcolorbox{Forecast}[1][]{
  basecolorbox,
  colback=forecastblue!10,
  colframe=forecastblue!90,
  title={#1}
}

\newtcolorbox{Contrast}[1][]{
  basecolorbox,
  colback=contrastterracotta!20,
  colframe=contrastterracotta!70,
  title={#1}
}

\newtcolorbox{AgreementMatrix}[1][]{
  basecolorbox,
  colback=agreementmustard!15,
  colframe=agreementmustard!80,
  title={#1}
}

\newtcolorbox{Crux}[1][]{
  basecolorbox,
  colback=cruxgreen!20,
  colframe=cruxgreen!70,
  title={#1}
}

\newtcolorbox{CodeBox}[1][]{
  codeboxstyle,
  colback=forecastblue!5,
  colframe=forecastblue!50,
  title={#1}
}

\usepackage{listings}
\usepackage{fvextra}  
\usepackage{multicol}  

\lstdefinestyle{pythonstyle}{
    language=Python,
    basicstyle=\ttfamily\small,
    keywordstyle=\color{forecastblue}\bfseries,
    stringstyle=\color{contrastterracotta},
    commentstyle=\color{cruxgreen},
    showstringspaces=false,
    breaklines=true,
    frame=none,
    xleftmargin=0pt
}

\usepackage{tikz}
\usetikzlibrary{positioning,fit,arrows.meta,calc}

\usepackage{enumitem}
\usepackage{thmtools}

\usepackage{amsmath,amsfonts,bm}

\def\eqref#1{equation~\ref{#1}}

\def\1{\bm{1}}

\DeclareMathAlphabet{\mathsfit}{\encodingdefault}{\sfdefault}{m}{sl}
\SetMathAlphabet{\mathsfit}{bold}{\encodingdefault}{\sfdefault}{bx}{n}

\definecolor{aricolor}{RGB}{200,0,120}
\definecolor{victorcolor}{RGB}{0,90,200}
\definecolor{davidcolor}{RGB}{0,110,50}
\definecolor{jamescolor}{RGB}{130,0,180}
\definecolor{claudecolor}{RGB}{200,90,0}
\definecolor{colmcolor}{RGB}{0,125,140}
\newif\ifcomments
\commentsfalse  

\title{Can LLMs Anticipate Their Cruxes?}

\author{
David Reber$^{1}$\thanks{Correspondence to \texttt{reber@uchicago.edu}.} \quad
James Reber$^{2}$ \quad
Ari Holtzman$^{1,3}$ \quad
Victor Veitch$^{3,4}$ \\[3pt]
$^{1}$Department of Computer Science, University of Chicago \quad
$^{2}$Independent Researcher \\
$^{3}$Data Science Institute, University of Chicago \quad
$^{4}$Department of Statistics, University of Chicago}

\iclrfinalcopy
\begin{document}

\maketitle
\ificlrfinal\hypersetup{pdftitle={Can LLMs Anticipate Their Cruxes?},
  pdfauthor={David Reber, James Reber, Ari Holtzman, Victor Veitch}}\fi
\ificlrfinal\lhead{Preprint. Under review.}\else\lhead{Under review as a conference paper at ICLR 2027}\fi

\vspace{-0.15in}
\begin{abstract}
   To use language models to help us make decisions we would like to understand the rationale behind the language model's outputs.
   In principle, the language model provides its rationale as part of its text output.
   However, the sheer volume of text makes this hard to use in practice.
   In this paper, we are interested in extracting the \emph{cruxes} that lead a model to recommend one action over another.
   We take a crux to be a sub-question that a model's rollouts disagree on, where changing its value may have altered the conclusion.
   A crux is therefore a property of the \emph{distribution} of a model's rollouts, not of any one rollout.
   We consider three strategies of increasing computational cost.
   First, we just ask a model to identify, a priori, what the cruxes of a decision will be.
   Second, we have the model actually go through the process of generating a response and then extract the cruxes that arose in its reasoning.
   Finally, we generate many responses and extract the cruxes by comparing the reasoning across the rollouts.
   We make this last procedure computationally tractable with cheap calls to a reranker, an estimated 86 times less compute than a judge.
   We then conduct an extensive empirical evaluation to assess whether the computationally inexpensive options are sufficient.
   They are not. Across domains and models, we find that a model's a priori assessment of the cruxes has little relation to the factors that are actually identified as important a posteriori.
   Using a single rollout helps, but is still inadequate for identifying the full set of cruxes.
   Accordingly, for high-stakes decisions (and to understand the space of language model decision making) the many rollout procedure is unavoidable.
\end{abstract}

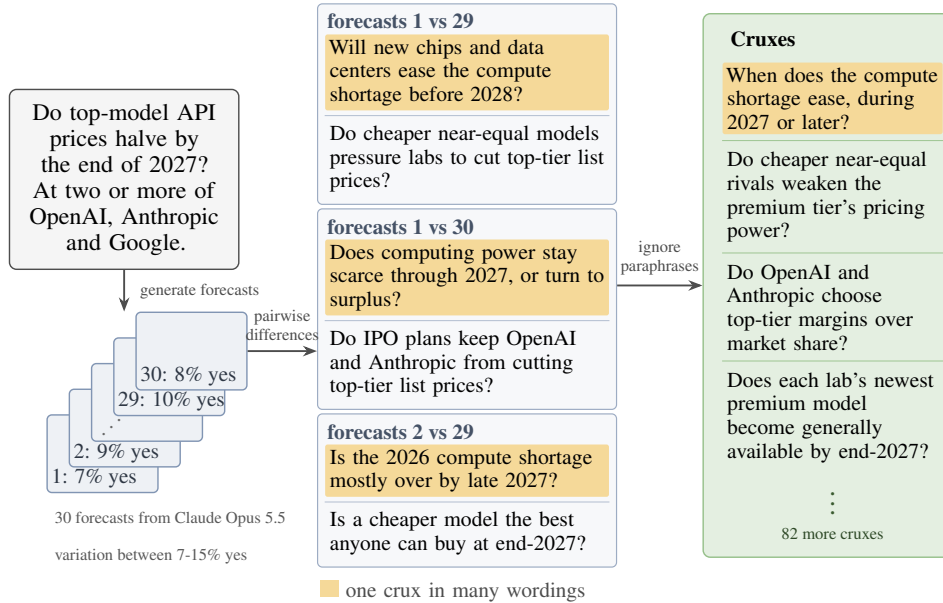
\begin{figure}[H]
\centering
\usetikzlibrary{backgrounds}
\definecolor{samecrux}{RGB}{230,159,0}
\newcommand{\figaccent}[1]{\sbox0{\textcolor{cruxgreen!45!black}{#1}}\rlap{\usebox0}\raisebox{-3pt}{\textcolor{cruxgreen!70}{\rule{\wd0}{0.8pt}}}}
\resizebox{0.90\linewidth}{!}{%
\begin{tikzpicture}[
  every node/.style={font=\small},
  qbox/.style={draw=black!70, line width=0.6pt, rounded corners=3pt, fill=black!4,    text width=1.1745in, align=center, inner sep=6pt, font=\normalsize, execute at begin node={\hyphenpenalty10000\exhyphenpenalty10000\relax}},
  card/.style={draw=forecastblue!90, line width=0.6pt, rounded corners=2pt,    fill=forecastblue!12, minimum width=0.6480in, minimum height=0.4536in},
  clabel/.style={font=\footnotesize, text=forecastblue!40!black, inner sep=2pt, anchor=south west},
  pline/.style={font=\footnotesize, text width=1.6038in, align=left, inner sep=2pt,    rounded corners=1pt, execute at begin node={\hyphenpenalty10000\exhyphenpenalty10000\relax}},
  same/.style={fill=samecrux!40},
  ptitle/.style={font=\footnotesize\bfseries, text=forecastblue!60!black, inner sep=1pt, anchor=west},
  pbox/.style={draw=forecastblue!70, line width=0.5pt, rounded corners=2pt, fill=forecastblue!6, inner sep=3pt},
  cline/.style={font=\small, text width=1.2312in, align=left, inner sep=2pt,    rounded corners=1pt, execute at begin node={\hyphenpenalty10000\exhyphenpenalty10000\relax}},
  cbox/.style={draw=cruxgreen!70, line width=0.6pt, rounded corners=3pt, fill=cruxgreen!18, inner sep=4pt},
  arr/.style={-{Stealth[length=6pt]}, line width=0.8pt, black!70}
]
\coordinate (px) at (1.8468in,0);
\node[ptitle] (t1) at ($(px)+(0,1.5228in)$) {forecasts 1 vs 29};
\node[pline, same, anchor=north west] (a1) at ($(t1.south west)+(0,-1pt)$) {Will new chips and data centers ease the compute shortage before 2028?};
\node[pline, anchor=north west] (b1) at ($(a1.south west)+(0,-4pt)$) {Do cheaper near-equal models pressure labs to cut top-tier list prices?};
\node[ptitle] (t2) at ($(b1.south west)+(0,-0.17in)$) {forecasts 1 vs 30};
\node[pline, same, anchor=north west] (a2) at ($(t2.south west)+(0,-1pt)$) {Does computing power stay scarce through 2027, or turn to surplus?};
\node[pline, anchor=north west] (b2) at ($(a2.south west)+(0,-4pt)$) {Do IPO plans keep OpenAI and Anthropic from cutting top-tier list prices?};
\node[ptitle] (t3) at ($(b2.south west)+(0,-0.17in)$) {forecasts 2 vs 29};
\node[pline, same, anchor=north west] (a3) at ($(t3.south west)+(0,-1pt)$) {Is the 2026 compute shortage mostly over by late 2027?};
\node[pline, anchor=north west] (b3) at ($(a3.south west)+(0,-4pt)$) {Is a cheaper model the best anyone can buy at end-2027?};
\coordinate (mid) at ($(t1.north)!0.5!(b3.south)$);
\coordinate (lx) at (0,0);
\coordinate (rx) at (4.1148in,0);
\coordinate (qtop) at ($(lx |- mid)+(0,1.15in)$);
\node[qbox, anchor=north west] (q) at (qtop) {Do top-model API prices halve by the end of 2027?\\At two or more of OpenAI, Anthropic and Google.};
\node[card, anchor=north west] (c1) at ($(q.south west)+(0.22in,-0.2430in-0.60in)$) {};
\node[card, anchor=north west] (c2) at ($(c1.north west)+(0.13in,0.15in)$) {};
\node[card, anchor=north west] (c3) at ($(c2.north west)+(0.13in,0.15in)$) {};
\node[card, anchor=north west] (c4) at ($(c3.north west)+(0.13in,0.15in)$) {};
\node[card, anchor=north west] (c5) at ($(c4.north west)+(0.13in,0.15in)$) {};
\node[clabel] at (c1.south west) {1: 7\% yes};
\node[clabel] at (c2.south west) {2: 9\% yes};
\node[font=\small, text=black!60, inner sep=0pt, anchor=south west] at ($(c3.south west)+(0.05in,0.025in)$) {\reflectbox{\ensuremath{\ddots}}};
\node[clabel] at (c4.south west) {29: 10\% yes};
\node[clabel] at (c5.south west) {30: 8\% yes};
\node[font=\scriptsize, text=black!65, anchor=north west] (mf) at ($(c1.south west)+(0,-0.06in)$) {30 forecasts from Claude Opus 5.5};
\node[font=\scriptsize, text=black!65, anchor=north west] (mf1) at ($(mf.south west)+(0,-0.03in)$) {variation between 7-15\% yes};
\coordinate (xtop) at ($(rx |- mid)+(0,1.4403in)$);
\node[font=\small\bfseries, anchor=west] (xt) at (xtop) {\hspace{4pt}Cruxes};
\node[cline, same, anchor=north west] (x1) at ($(xt.south west)+(4pt,-3pt)$) {When does the compute shortage ease, during 2027 or later?};
\node[cline, anchor=north west] (x2) at ($(x1.south west)+(0,-5pt)$) {Do cheaper near-equal rivals weaken the premium tier's pricing power?};
\node[cline, anchor=north west] (x3) at ($(x2.south west)+(0,-5pt)$) {Do OpenAI and Anthropic choose top-tier margins over market share?};
\node[cline, anchor=north west] (x4) at ($(x3.south west)+(0,-5pt)$) {Does each lab's newest premium model become generally available by end-2027?};
\node[font=\large, text=cruxgreen!55!black, anchor=north] (vd) at ($(x4.south west)!0.5!(x4.south east)+(0,-1pt)$) {$\vdots$};
\node[font=\scriptsize, text=cruxgreen!45!black, anchor=north] (vdl) at ($(vd.south)+(0,0.02in)$) {82 more cruxes};
\coordinate (hcx) at ($(q.west)!0.5!(x4.east)$);
\coordinate (htop) at ($(hcx |- t1.north)+(0,0.1620in)$);
\node[font=\normalsize\bfseries, anchor=south] (head) at (htop) {Cruxes are properties of the \figaccent{distribution} of forecasts, not of any one forecast.};
\begin{scope}[on background layer]
\node[pbox, fit=(t1)(a1)(b1)] (p1) {};
\node[pbox, fit=(t2)(a2)(b2)] (p2) {};
\node[pbox, fit=(t3)(a3)(b3)] (p3) {};
\node[cbox, fit=(xt)(x1)(x2)(x3)(x4)(vd)(vdl)] (x) {};
\end{scope}
\foreach \u/\l in {a1/b1,a2/b2,a3/b3}{\draw[forecastblue!55, line width=0.4pt] ($(\u.south west)!0.5!(\l.north west)$) -- ($(\u.south east)!0.5!(\l.north east)$);}
\foreach \u/\l in {x1/x2,x2/x3,x3/x4}{\draw[cruxgreen!65, line width=0.4pt] ($(\u.south west)!0.5!(\l.north west)$) -- ($(\u.south east)!0.5!(\l.north east)$);}
\draw[arr] (q.south) -- (q.south|-c5.north) node[midway, right, font=\scriptsize, text=black!70, align=left, xshift=3pt] {generate forecasts};
\draw[arr] (c5.east) -- (p2.west|-c5.east) node[midway, above, font=\scriptsize, text=black!70, align=center, yshift=1pt] {pairwise\\differences};
\draw[arr] (p2.east|-x.west) -- (x.west) node[midway, above, font=\scriptsize, text=black!70, align=center, yshift=1pt] {ignore\\paraphrases};
\node[font=\footnotesize, text=black!70, anchor=north west, text width=2.19in, align=left, inner sep=1pt] (lg) at ($(p3.south west)+(0,-4pt)$) {\tikz\fill[samecrux!40] (0,0) rectangle (8pt,7pt); one crux in many wordings};
\end{tikzpicture}
}
\caption{A crux is an assumption that forecasts disagree on, and whose answer could have changed their conclusions.}
\label{fig:overview}
\end{figure}

\section{Introduction}\label{sec:introduction}

Consider using a language model to aid in decision making.
For example, someone asking a chatbot whether to end a relationship, a patient weighing an AI-assisted diagnosis, or a researcher considering an agent-implemented change to their codebase.
To decide what the right decision is, we would like to reduce the overall argument to the critical subquestions that lead to the conclusion.
These subquestions are the \emph{cruxes} of the argument \citep{sabien2018doublecrux}.
They should surface the assumptions and reasoning that influence the conclusion.
For example, suppose a coding agent skips every input that crashes its pipeline.
The agent has decided, without saying so, that the skipped inputs do not matter.
Whether the skipped inputs matter is a crux of the researcher's decision to accept the change.
How might the researcher identify this implicit crux if the agent does not volunteer it?

The most obvious option is to simply ask the language model what the cruxes are.
We can do this a priori---just ask in advance what the relevant subquestions will be---or a posteriori---have the language model generate a decision, and then look at its chain of thought to determine what the key subquestions were.
The a priori case matters because cruxes named in advance say what to find out before deciding.
For example, a priori cruxes let a language model ask for clarification before carrying out a task.
The a posteriori case would hopefully be useful to sanity check the model's reasoning.
Both the a priori and single-rollout methods are appealing for their simplicity and inexpensiveness.
Is either of these adequate?

\begin{figure}[t!]
\centering
\begin{minipage}[t]{0.46\linewidth}
\centering
\includegraphics[width=0.85\linewidth]{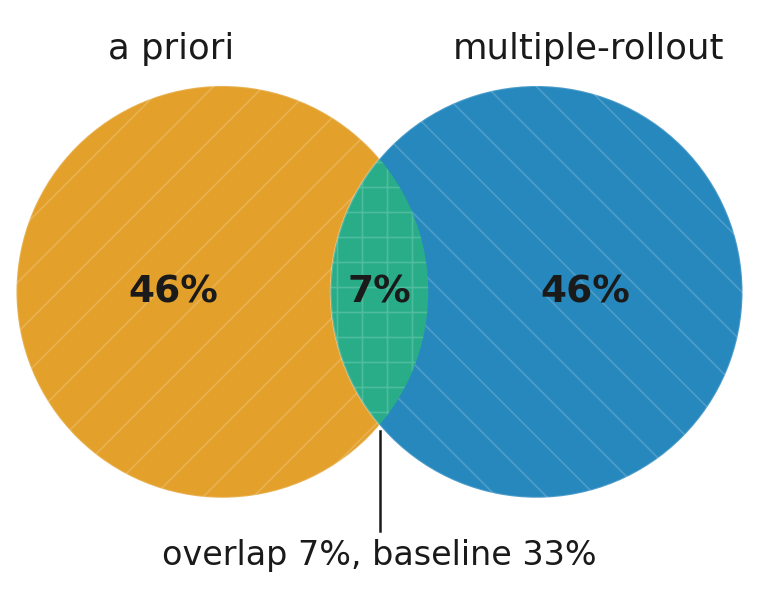}

\small
\textbf{A priori:} ``Does free-throw performance become a decisive factor?''

\textbf{Multiple-rollout, a crux question as one forecast answered it:} ``Iowa State forced 23 turnovers against Arizona State on March 7.''
\end{minipage}
\begin{minipage}[t]{0.46\linewidth}
\centering
\includegraphics[width=0.85\linewidth]{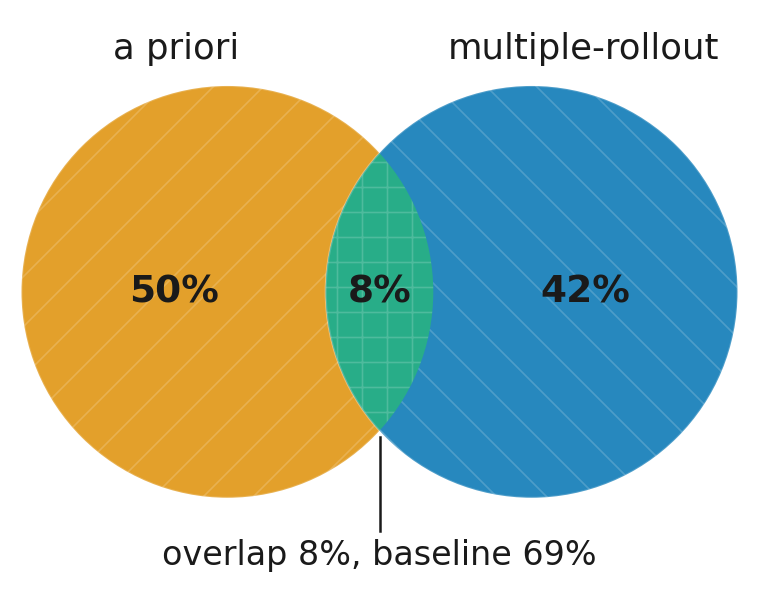}

\small
\textbf{A priori:} ``Will Frontier maintain sufficient unrestricted liquidity and borrowing capacity through April 2026 to avoid a distress filing?''

\textbf{Multiple-rollout, a crux question as one forecast answered it:} ``\mbox{Altman Z-score $\approx 0.38$}: distress zone.''
\end{minipage}
\caption{%
In a basketball game and a bankruptcy question, the cruxes a model predicted matched the questions its forecasts varied on far less often than the permutation baseline, which reassigns the same cruxes at random to the a priori and multiple-rollout sets.
\textbf{Left:} Iowa State vs.\ Kentucky, a game in March Madness, the American college basketball tournament.
\textbf{Right:} Frontier Group bankruptcy, an event on Metaculus, a public forecasting platform.
Each diagram splits the clusters of paraphrased cruxes (\cref{sec:equivalence}), formed from 25 cruxes per set, into those holding a priori cruxes only (orange), both kinds (green, the overlap) and multiple-rollout cruxes only (blue).
Below each diagram stand one a priori crux, as a question, and one multiple-rollout crux, as one forecast's answer (\textbf{contribution 1} extracts the multiple-rollout cruxes; \textbf{contribution 2} groups similar paraphrases at low cost).
A turnover is a possession lost before the team can shoot, and a forced turnover is one the defense causes.
The Altman Z-score is a bankruptcy-risk score computed from a company's accounts.
Below the circles each diagram prints its overlap at 25 cruxes and its permutation baseline (\textbf{contribution 3} runs this comparison across games, events and models).
}
\label{fig:composite_venns}
\end{figure}

To study this question, we introduce a method for extracting cruxes by contrasting many rollouts aimed at the same decision (\cref{fig:overview}, whose run is described in \cref{app:fig1demo}).
Intuitively, the idea is that if a question is a crux of the reasoning then we should actually see variation across independent reasoning of the decision!
To make this concrete, consider the forecasting question of whether Frontier Airlines files for bankruptcy.
A priori, the bot identifies ``whether Frontier keeps enough cash to avoid a distress filing'' as an important factor.
Given a single rollout, the bot might actually end up focusing on ``whether Frontier has enough cash to meet its obligations'', adding \emph{another} crux.
Given multiple rollouts, if the bot always had the same belief about Frontier's cash position, that belief would not be a crux of the reasoning!
But if the rollouts vary in how much credence they give to a rival airline's recent bankruptcy, that \emph{is} a crux (\cref{fig:composite_venns}, right).
The question is whether we really need the multiple rollouts, or whether LLMs are already capable of identifying, a priori or after a single rollout, the cruxes of their own reasoning.
To address this:
\begin{enumerate}\setlength{\itemsep}{0pt}\setlength{\parsep}{0pt}
   \item We propose an elegant (but naively, expensive!) method for extracting the cruxes of a model's conclusion from multiple rollouts.
   Informally, the idea is to contrast each pair of rollouts looking for important distinctions in the reasoning.
   Equivalent paraphrases of the same crux are then clustered.
   \item We make this multiple-rollout method affordable.
   A reranker, which scores a pair of cruxes in a single forward pass, replaces an LLM judge for deciding whether two cruxes are equivalent.
   It matches every pair of the main evaluation in 6.4 GPU-hours, an estimated 86 times less compute than a local judge, and still agrees with it on most pairs and rarely merges two cruxes it called distinct (\cref{sec:equivalence,app:costs}).
   \item We evaluate the a priori and single-rollout methods against the multiple-rollout method.
   A method could score well merely by naming more cruxes, or by naming one crux many ways, so the overlap between two sets is read against a permutation baseline, which reshuffles the same labels over the same clusters (\cref{sec:exp_headline}).
   The comparison runs on two sources of forecasting problems, four model families and three definitions of semantic similarity.
\end{enumerate}
Simply asking for the cruxes is plainly inadequate; the models do worse than a random shuffle (\cref{fig:composite_venns}).
Further, using only a single rollout typically misses many cruxes.
Accordingly, for high-stakes decisions, the multiple-rollout approach is necessary.

\section{Related Work}\label{sec:related}

Three lines of work come closest to this paper: introspection, claim extraction, and forecasting systems.
Introspection asks if a model can report on itself: whether it knows what it knows~\citep{kadavath2022language,lin2022teaching,kapoor2024taught}, whether it can describe its own tendencies~\citep{binder2025looking,betley2025tell,laine2024sad,lindsey2026emergent,comsa2025does}, and whether its written reasoning is faithful to what produced its answer~\citep{wei2022chain,turpin2024language,chen2025reasoningmodelsdontsay,chua2025deepseekr1reasoningmodels}.
None of that work asks whether a model can say in advance which factual questions its own forecasts will disagree on.
Claim extraction breaks a text into claims~\citep{barhaim2020kpa,min2023factscore,wei2024safe}, and self-consistency uses the spread among many outputs to reach a better answer~\citep{wang2023selfconsistency,du2024improving}.
The comparison here instead reads two forecasts at once and keeps only the claims on which they differ.
Forecasting systems search for evidence before they forecast~\citep{halawi2024approaching,murphy2026agentic}, and the Metaculus forecasts here come from a system of that kind.
That work aims to make each forecast better; this paper takes the forecasts as given and asks how to find their cruxes.
CruxBench, a concurrent benchmark, scores the crux a model names by how far its resolution moves a prediction market~\citep{dai2026cruxbenchbenchmarkinformationdiscovery}; the test here asks instead whether the cruxes a model names are the ones its own forecasts disagree on.
In CruxBench the strongest model scores 0.016 bits, narrowly above the 0.007 of a baseline that picks the resolution time at random.
Each line is described in greater detail in \cref{app:related}.

\section{Data Sources}\label{sec:forecasting}

A particularly clean task with high impact on decision-making is forecasting, where a model outputs a probability for a future event with supporting rationale.
For our purposes, we restrict to binary events, where the model must choose between two mutually exclusive outcomes; for example, a model might forecast the probability that Frontier Airlines files for bankruptcy before May 2026.

A forecasting agent produces each forecast in three steps.
First, a deep research assistant searches the web.
Second, the agent picks the factors that matter most.
Third, the agent weighs the evidence on each factor and writes a probability for the event.

All forecasts were made live, between 2026-02-28 and 2026-03-18, before the events resolved.
The multiple-rollout procedure reads cruxes off the disagreements between rollouts, so what a disagreement means depends on what makes two rollouts differ.
The detail to keep in mind about each data source is therefore where the variation among its rollouts comes from.

\subsection{Metaculus}\label{sec:metaculus}
Metaculus is a public forecasting platform that hosts tournaments on a wide range of topics: we consider 15 Metaculus events, spanning topics from geopolitics to economics to science and technology (\cref{app:metaculus}).

Each of the 15 events received 30 rollouts.
One of the 450 rollouts was erroneously not stored, and a rollout made after the events resolved would no longer be live, so the missing one was not rerun; that event has 29 rollouts and the others 30.
Each rollout received the same prompt but ran its own web searches, six calls that gather background and then three probability estimates from different framings, with no search result shared across rollouts.

Metaculus rollouts differ through the model's sampling and through the web search results each received.
Its rollouts are written by gpt-5-nano with web search, and the cruxes predicted in advance come from gpt-5.4 and those read off the rollouts from gpt-5.4-mini (\cref{tab:roles}), so its 15 events give breadth of topic with one model family.

\subsection{March Madness}\label{sec:marchmadness}
March Madness is a single-elimination basketball tournament run each year by the National Collegiate Athletic Association in the United States.
The experiments involve predicting the outcome of 59 of its games, each between two teams, with every forecast made on 2026-03-17 and 03-18, before the tournament began.
March Madness may not be a high-stakes setting, but it suits a study of the three procedures because its evidence can be held fixed.
A Metaculus question depends on the latest news, whereas the evidence about a basketball team can be gathered once, before the first game, and every rollout of every model then reasons from the same evidence.
The variation that remains is the model's own, and that is what a comparison across models needs.

The same forecasting agent as above was used to forecast the winner of each game.
Since there were 1{,}891 possible matchups among the 62 teams, the agent's web-search stage ran once per team rather than once per rollout.
It wrote three scouting reports on each team, which were cached and reused for every potential matchup.
The rollouts on a game therefore read those reports instead of searching the web for the same information.
This lets us evaluate the conditional generative distribution of each model more narrowly.

Each report was written from one of three personas to induce different perspectives on the same team.
The personas were a `Traditional Basketball Analyst', a `Quantitative Forecaster', and a `Matchup Analyst'.

Hence, in contrast to Metaculus, no variation enters through web search, by construction, so the comparison can focus on the models.
In the main runs gpt-5.4-nano writes the rollouts, predicts the a priori cruxes, and reads the cruxes off a single rollout and off many rollouts.
The same chain was run on each of three open-weight models, qwen3:32b, gemma4:31b and gpt-oss:120b (\cref{sec:exp_models}).
Each rollout reads one of the three reports on each team, so the $3 \times 3$ pairs of reports give 9 rollouts per game.

\begin{figure}[t!]
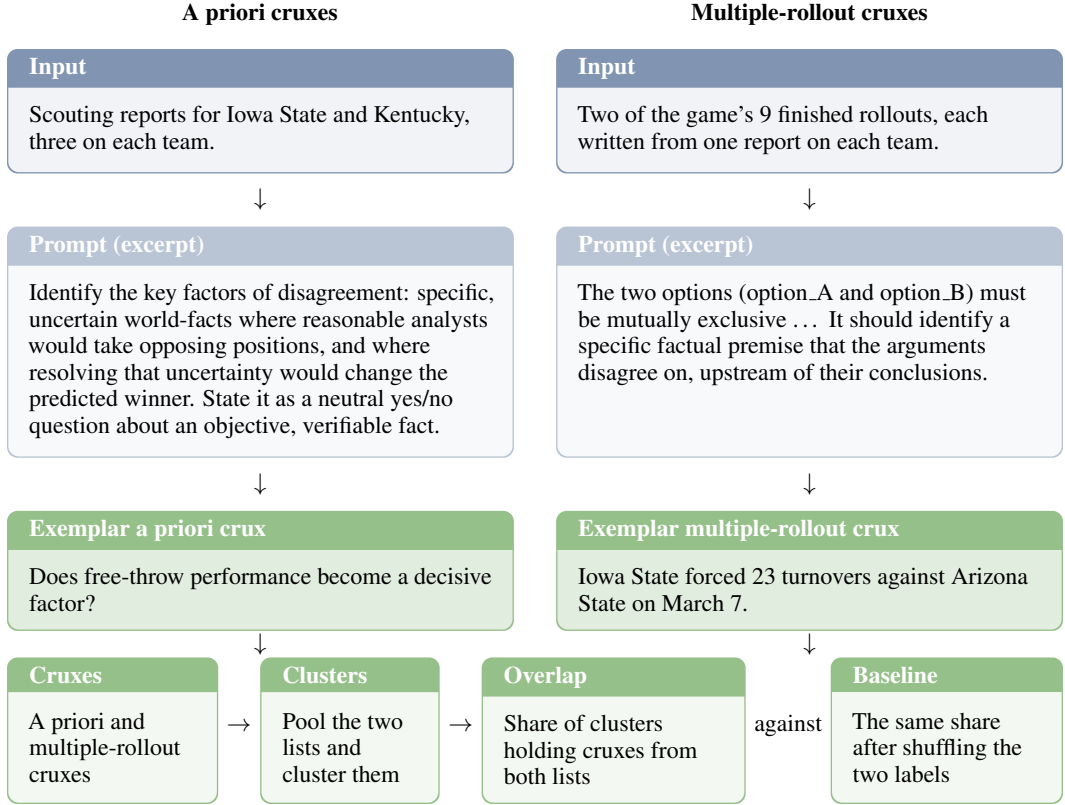

\centering
\small
\tcbset{promptfig/.style={enhanced, boxrule=0.5pt, arc=4pt, outer arc=4pt, left=5pt, right=5pt, top=4pt, bottom=4pt, fonttitle=\bfseries, halign=flush left}}
\newcommand{\promptcol}[5]{%
\begin{minipage}[t]{0.48\linewidth}
\centering\textbf{#1}\par\smallskip
\begin{tcolorbox}[promptfig, colback=forecastblue!10, colframe=forecastblue!90, title=Input, equal height group=promptin]#2\end{tcolorbox}
$\downarrow$
\begin{tcolorbox}[promptfig, colback=forecastblue!5, colframe=forecastblue!50, title={Prompt (excerpt)}, equal height group=promptask]#3\end{tcolorbox}
$\downarrow$
\begin{tcolorbox}[promptfig, colback=cruxgreen!20, colframe=cruxgreen!70, title={#4}, equal height group=promptout]#5\end{tcolorbox}
\end{minipage}}
\promptcol{A priori cruxes}
  {Scouting reports for Iowa State and Kentucky, three on each team.}
  {Identify the key factors of disagreement: specific, uncertain world-facts where reasonable analysts would take opposing positions, and where resolving that uncertainty would change the predicted winner. State it as a neutral yes/no question about an objective, verifiable fact.}
  {Exemplar a priori crux}
  {Does free-throw performance become a decisive factor?}
\hfill
\promptcol{Multiple-rollout cruxes}
  {Two of the game's 9 finished rollouts, each written from one report on each team.}
  {The two options (option\_A and option\_B) must be mutually exclusive \ldots{} It should identify a specific factual premise that the arguments disagree on, upstream of their conclusions.}
  {Exemplar multiple-rollout crux}
  {Iowa State forced 23 turnovers against Arizona State on March 7.}

\begin{minipage}{0.48\linewidth}\centering$\downarrow$\end{minipage}\hfill\begin{minipage}{0.48\linewidth}\centering$\downarrow$\end{minipage}

\tcbset{pipefig/.style={promptfig, colback=cruxgreen!8, colframe=cruxgreen!70, nobeforeafter, box align=center, valign=center, equal height group=pipefig}}
\begin{tcolorbox}[pipefig, width=0.2\linewidth, title=Cruxes]A priori and multiple-rollout cruxes\end{tcolorbox}%
\hfill$\rightarrow$\hfill
\begin{tcolorbox}[pipefig, width=0.17\linewidth, title=Clusters]Pool the two lists and cluster them\end{tcolorbox}%
\hfill$\rightarrow$\hfill
\begin{tcolorbox}[pipefig, width=0.25\linewidth, title=Overlap]Share of clusters holding cruxes from both lists\end{tcolorbox}%
\hfill against\hfill
\begin{tcolorbox}[pipefig, width=0.22\linewidth, title=Baseline]The same share after shuffling the two labels\end{tcolorbox}
\caption{In March Madness the test asks whether a model, reading every scouting report its forecasts were written from, can foresee where those forecasts will disagree.
A priori cruxes (left) against multiple-rollout cruxes (right), for one March Madness game, Iowa State vs.\ Kentucky.
The a priori prompt predicts cruxes from the reports, and the multiple-rollout prompt finds them in each pair of the game's rollouts.
The a priori and multiple-rollout cruxes are then pooled and clustered, and their overlap is compared with a baseline that shuffles the two labels among the same cruxes (bottom row).
Each crux is a yes/no question with both answers, and the figure quotes the multiple-rollout crux by the assumption one of the two rollouts took, that Iowa State forced 23 turnovers in an earlier game.}
\label{fig:prompts}
\end{figure}

\section{Crux Extraction Methods}\label{sec:methods}

The question this paper asks is: Given a fixed forecaster, which method of crux extraction is best?
The first method asks the model in advance what the cruxes of its decision will be (a priori).
The second has the model generate one rollout and then extracts the cruxes from that single rollout (single-rollout).
The third method generates many rollouts and extracts the cruxes by comparing pairs of them (multiple-rollout).
\Cref{fig:prompts} shows the a priori and multiple-rollout prompts for one game.
For the experiment, the forecaster is better considered as part of the data, because in a high-stakes setting one would expect one model to generate the rollout and make the recommendation while a different model evaluates its trustworthiness.
\Cref{tab:roles} lists the model in each role, forecaster, predictor and extractor, and any model can play any part (\cref{sec:exp_models}).

\paragraph{Predicting the cruxes in advance (a priori).}\label{sec:rationales}\label{sec:asking} The a priori method asks the model, without seeing any rollout, on which cruxes the rollouts will take different sides.
In Metaculus the model receives only the question's text, with no background, forecasts or web search results, so the rollouts' web searches may find evidence it lacks (\cref{app:designs} weighs what this can explain), and it is sent 11 prompts, each 5 times, for 55 calls per event (\cref{app:costs}).
In March Madness the model, the same one that writes the rollouts, receives all three scouting reports on each team, the same reports the game's rollouts are written from.
Five differently worded prompts are sent once each, and every crux they return joins the a priori set, with repeats removed (\cref{sec:exp_headline}).

\paragraph{Extracting the cruxes from a single rollout (single-rollout).}\label{sec:oneshot} The single-rollout method picks one rollout per game or event at random, with a fixed seed.
A prompt written for one rollout then extracts the cruxes of that rollout alone, about 12 per game and 6 per event (\cref{app:single}).
Its extraction prompt is intentionally kept as close as possible to the multiple-rollout extraction prompt, differing only where one rollout replaces a pair (\cref{tab:prompt_diff}).
\Cref{app:rollout} shows one Metaculus rollout in full.

\paragraph{Discovering the cruxes from many rollouts (multiple-rollout).}\label{sec:cruxes} This method of extracting cruxes instead contrasts two rollouts on the same request and asks the extractor for the factual questions on which the two take different sides.
Note that this has the advantage of not having to enumerate all possible factors that could be important; by construction it restricts to the factors that are relevant to the model's sampling distribution.
Taking rollouts two at a time ties each crux to one pair of rollouts that disagree on it and records which side each of the two took.
Whichever method produced it, a crux is stored as a yes/no question together with its two answers, one for each side a forecast could take; examples in this paper often quote a crux by one side's answer, such as ``world demand for oil will have peaked in ten years''.
\Cref{fig:prompts} shows both prompts for one March Madness game.

\begin{figure}[t]
\centering
\includegraphics[width=\linewidth]{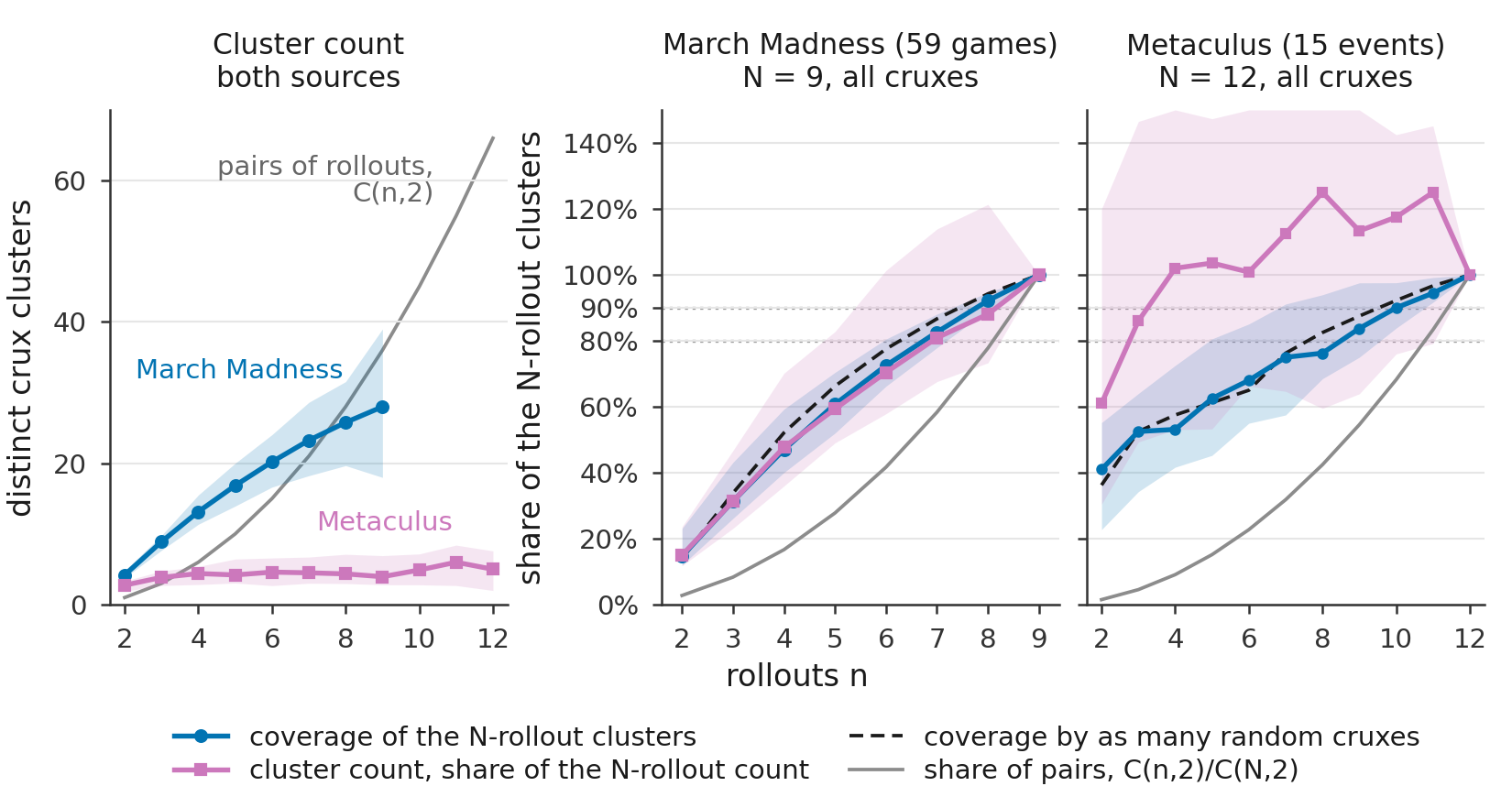}
\caption{Cruxes are often shared across rollouts, so the number of distinct cruxes grows about linearly with the number of rollouts even though the number of rollout pairs grows quadratically.
\textbf{Left:} the median number of distinct crux clusters at the same-resolution level (\cref{sec:equivalence}) against the number of rollouts, each game or event averaged over up to 20 random subsets (bands, 10th to 90th percentile), with the number of rollout pairs for reference.
March Madness reaches 28 clusters at nine rollouts and shows no sign of saturation.
Metaculus, with 12 of the 29 or 30 rollouts per event and all their cruxes, appears to converge to 5 cruxes.
\textbf{Middle and right:} March Madness (59 games, $N = 9$) and Metaculus (15 events, $N = 12$), where the colours now mark quantities, not sources, and every curve is a share of what all $N$ rollouts give.
Blue is coverage, the share of the clusters from all $N$ rollouts that $n$ rollouts already find.
The dashed line is the coverage of as many cruxes drawn at random from the pool.
Pink is the number of clusters from $n$ rollouts as a share of the number from all $N$.
In Metaculus that ratio passes 100\%, and the axis stops at 150\%.}
\label{fig:recurrence}
\end{figure}

Every pair of rollouts is contrasted, even two that reach the same conclusion, because they can still take different sides on a crux (\cref{sec:exp_variation}).
For example, both rollouts can back the oil bet, with one expecting world demand to keep growing and the other expecting it to peak while supply shrinks faster.
Even though $n$ rollouts give $\binom{n}{2}$ pairs to contrast, they produce far fewer distinct cruxes, because the cruxes repeat heavily across pairs: once paraphrases of one crux are grouped, the number of distinct cruxes grows about linearly with the number of rollouts, about 3.4 per rollout in March Madness, from 4 at two rollouts to 28 at nine, while the number of pairs reaches 36 (\cref{fig:recurrence}, left).
Because the method starts from disagreements the model actually produced, it never has to enumerate every factor a forecast could depend on.
The cruxes it finds also recur across rollouts.
Five of nine rollouts give 28\% of the pairs yet 61\% of the distinct cruxes all nine give, and seven give 83\% (\cref{fig:recurrence}, middle).
In Metaculus each event is scored on twelve rollouts, and a subset of those twelve already finds most of the distinct cruxes all twelve give (\cref{fig:recurrence}, right).
Some subsets even give more distinct cruxes than all twelve together.

\section{Evaluation Methodology}\label{sec:eval}
In order to evaluate the three methods of crux extraction, we need a way to compare the cruxes they produce.
There are two key challenges in doing this.
First, the cruxes are expressed in natural language, which means that two cruxes that are semantically equivalent may be phrased very differently.
Second, we need a way to quantify the overlap between two sets of cruxes in a way that is invariant to how many each method produces.
To address these challenges, we introduce a method for clustering cruxes based on their semantic similarity and then measuring the overlap between the resulting clusters.
\subsection{Equivalence between Cruxes}\label{sec:equivalence}

Because a natural-language crux can be phrased many different ways, how are we to judge whether two cruxes are equivalent?

Two cruxes are treated as the same when the same evidence would confirm or falsify both.\footnote{A match does not require an a priori crux to predict which side any rollout took.
One set of cruxes may ask whether world demand for oil will have peaked in ten years and the other whether it will still be growing, so the first answer of one crux is the second answer of the other.
Both sets label a crux's two answers A and B, and the instruction behind every reported number tells the reranker, introduced below, to ignore their order (\cref{app:robustness}).}
A generative judge cannot score every pair.
With $n$ rollouts and $\kappa$ cruxes returned per contrasted pair, a naive all-pairs comparison needs
\begin{equation}
K = \kappa\binom{n}{2} = \Theta\!\left(\kappa n^{2}\right),
\qquad
\binom{K}{2} = \frac{\kappa^{2} n^{2} (n-1)^{2}}{8} - \frac{\kappa\,n(n-1)}{4} = \Theta\!\left(\kappa^{2} n^{4}\right),
\label{eq:naive}
\end{equation}
Averaged over games or events, each with its own $K$, that is 12{,}978 judgements per March Madness game ($n = 9$, mean $\kappa = 4.47$) and 1{,}013{,}888 per Metaculus event ($n = 29$ or $30$, mean $\kappa = 3.27$) at one level of strictness.

Two choices make the comparison affordable.
First, Qwen3-Reranker-8B~\citep{qwen3embedding} scores each pair in place of a generative judge.
It scores whether one passage matches another under a written instruction, in one forward pass of about 270 tokens.
The method's cost therefore splits into generations and forward-pass scores,
\begin{equation}
\underbrace{n + \binom{n}{2}}_{\text{generations, } \Theta(n^{2})}
\;+\;
\underbrace{\binom{K}{2}}_{\text{reranker passes, } \Theta(\kappa^{2}n^{4})}
\qquad
P = \binom{2m}{2} = 1{,}225 \ \text{pairs at } m = 25,
\label{eq:pipeline}
\end{equation}
and every pair of the complete March Madness multiple-rollout cruxes was scored once, for \cref{fig:recurrence}.
Those 751{,}515 pairs, at the same-resolution level defined below, took 12 GPU-hours.
A reranker score can change when the two cruxes swap places, so the similarity is the mean of both passes, each crux in turn as the query, $s(i,j) = \tfrac{1}{2}\big(r(i,j) + r(j,i)\big)$.
The reranker and the clustering below were checked against 2,000 March Madness crux pairs judged by gpt-5.4-nano, and they agree with it at a small fraction of a judge's cost (\cref{app:validation}).

\begin{figure}[tp]
\centering
\includegraphics[width=\linewidth]{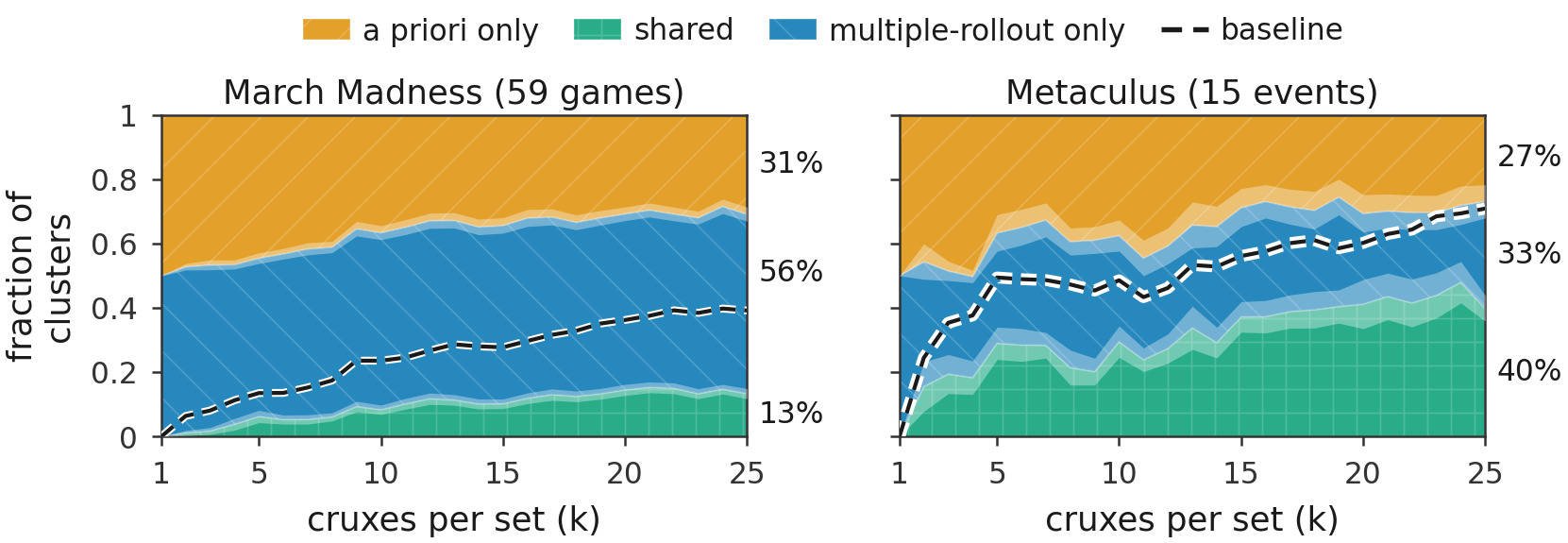}
\caption{Naming more cruxes, up to 25 on each side, still does not close the gap between predicted and discovered cruxes.
Bands give the share of clusters holding a priori cruxes only (top, orange), multiple-rollout only (middle, blue) or both (bottom, green, the overlap).
The dashed line is the permutation baseline, the overlap when the two labels are shuffled.
\textbf{Left:} March Madness, 59 games.
\textbf{Right:} Metaculus, 15 events, baseline 49.5\% at 5 cruxes and 70.9\% at 25.
Same-resolution matching throughout; lighter bands show $\pm 1$ SE, and the numbers at the right give each band's share at 25 cruxes.}
\label{fig:stacked}
\end{figure}

Second, for the overlap comparison each set is capped at $m = 25$ cruxes by a seeded draw.
The pooled $2m$ cruxes then give the $P$ pairs of \cref{eq:pipeline} per question and per level.
The cap fixes the cost; it does not make the full comparison cheaper.
The ratio $\binom{K}{2}/P$ of at least 10.4 in March Madness and 808 in Metaculus is a lower bound on what the cap saves (\cref{app:costs}).
The sweep of set sizes in \cref{fig:stacked} shows the gap at every size up to 25, so the result does not depend on the cap.
At each size $k$, each set gives $k$ of its cruxes, or all of them when it holds fewer.
In Metaculus the size of the gap moves by about ten points on a second draw of the 25, but the gap never closes (\cref{app:alignment}).

The rule leaves open how much two cruxes must share, so the reranker scores each pair at three levels of strictness.
At the strict level, \emph{same question}, two cruxes match when they address the same underlying point of disagreement, however they are worded.
At the next level, \emph{same resolution}, two cruxes match when resolving one would necessarily resolve the other, even if one is more specific.
At the broad level, \emph{same dispute}, two cruxes match when settling either would end the same disagreement between two rollouts.
\Cref{app:robustness} works through an oil example that separates the three levels.
The same-resolution level produces every number in the main text unless the text names another level.

Average-linkage hierarchical clustering groups the pooled cruxes at distance $1 - s(i,j)$, starting from single cruxes and merging the two groups whose mean pairwise distance is smallest until one group remains.
The order of those merges forms a tree, and cutting it stops the merges at a chosen number of clusters.
Persistence picks that number, the partition surviving the widest range of merge heights.
Ties go to fewer clusters, a single cluster is never chosen, and the cut never sees which set a crux came from.
At 25 cruxes per set the cut leaves on average 20.5 clusters per March Madness game (range 2 to 36).
The mean is 5.2 per Metaculus event (range 2 to 12).
Where the cut falls decides how different two cruxes can be and still share a cluster.
The comparison is therefore repeated in \cref{sec:exp_specificity} at cuts from broad categories down to one crux per cluster.

\subsection{Permutation Baseline to Address Singleton-Cluster Bias}\label{sec:exp_headline}\label{sec:experiments}

A cluster is \emph{shared} when it holds both a priori and multiple-rollout cruxes, and the obvious score is the fraction of clusters that are shared.
But this introduces yet another problem: many extracted cruxes are never marked equivalent to any other crux, so they sit alone in their own clusters.
A cluster of one can never be shared by the two sets, so the raw overlap is set as much by the cluster-size distribution as by agreement between the sets!

To correct for this singleton-cluster bias, the overlap is read against a permutation baseline, which holds the clusters fixed and reassigns the two labels at random.
The \emph{gap} is the baseline minus the observed overlap, in percentage points, so a gap near zero means the two sets cannot be told apart by their clusters.\footnote{Gaps are computed before rounding, so a gap can differ slightly from the difference of the rounded overlap and baseline.}
\Cref{app:metric} gives the baseline in closed form and defines the two further scores the paper quotes, coverage and the split-half check.

\section{Results}\label{sec:results}

Armed with this clustering and baseline, we can ask a more precise question:
\begin{quote}
Fix a level of equivalence and a number of cruxes on each side.
Do the a priori and single-rollout cruxes share clusters with the multiple-rollout cruxes more often than a shuffle of the same labels would?
\end{quote}

Each check recomputes the permutation baseline on the cruxes it scores, so the baselines differ from one check to the next.

\subsection{Predicted against Discovered Cruxes}\label{sec:res_headline}

The a priori cruxes match the multiple-rollout cruxes less often than a random reassignment of the same cruxes would.

In March Madness, at the same-resolution level with 25 cruxes on each side, the overlap averages 13.4\% $\pm$ 1.6\% (SE, the standard error across 59 games) against a permutation baseline averaging 39.2\%.
Growing both sets never lifts the overlap to the baseline (\cref{fig:stacked}).
Two halves of the multiple-rollout cruxes overlap 3.4 times as much as the two sets do (\cref{fig:withinset} in \cref{app:alignment}).
The gap below the baseline averages 25.9 $\pm$ 1.2 points ($p < 10^{-10}$), and 55 of the 59 games fall below their own baseline, three equal it and one lies above it (\cref{fig:per_matchup_hist}).

In March Madness the single-rollout cruxes come far closer to the baseline on overlap, 18.8\% against 25.1\%, below it in 47 of the 59 games (\cref{tab:single}).
That gap of 6.4 points is far smaller than the a priori cruxes' 26.8 on the same multiple-rollout cruxes.
The single-rollout cruxes still fall short on coverage, reaching 28.6\% of the multiple-rollout cruxes against a baseline of 37.8\% (\cref{tab:containment}).
In Metaculus the single-rollout cruxes overlap by 48.5\% against 44.0\% and sit at the baseline on every measure.
The single-rollout cruxes are therefore closer to a random sample of the multiple-rollout cruxes than the a priori cruxes are, and one sample is not the set.

\subsection{Sources and Contested Games}\label{sec:res_variation}\label{sec:exp_variation}

Could the gap come from a feature only one source has?
Each source is scored on its own.
A game whose rollouts all pick one team may seem to give the forecasts nothing to disagree about.
A March Madness game is \emph{contested} when its rollouts do not all pick the same winner.

\paragraph{Two sources.} In Metaculus the overlap is 39.9\% $\pm$ 4.0\% against a baseline of 70.9\%, a gap of 31.0 $\pm$ 4.9 points ($p = 1.2 \times 10^{-4}$), and 13 of the 15 events fall below their own baseline (\cref{app:alignment}).

\paragraph{Contested games.} The 29 contested games overlap 16.0\% $\pm$ 2.5\% against a baseline of 40.5\%, with 26 below their own baseline, one above it and two equal to it.

The gap therefore comes from neither the choice of source nor agreement on the winner.

\subsection{Strictness of Equivalence}\label{sec:res_strictness}

Does the gap depend on how strictly two cruxes must match to share a cluster?
The headline scores are repeated at the strict and broad levels of \cref{sec:equivalence}, in both sources.

At 25 cruxes the gap runs from 25.9 to 36.1 points across the three levels (\cref{tab:matching_criteria}; \cref{app:robustness}).
Neither a stricter nor a looser reading of ``same crux'' closes the gap in either source.

\subsection{Specificity and Coarser Clusters}\label{sec:res_specificity}\label{sec:exp_specificity}\label{sec:exp_specific}

Do the two sets miss each other because the a priori cruxes are vague rather than because they ask about different things?
Four checks cut the clustering coarser, ask the predictor for cruxes as specific as the multiple-rollout ones, rewrite the multiple-rollout cruxes vaguer, and match cruxes on the evidence behind them instead of their wording (\cref{app:alignment,app:specific}).

At every cut coarser than one crux per cluster the March Madness overlap stays below its baseline (\cref{fig:granularity}).
Asking for cruxes as specific as the multiple-rollout ones narrows the gap by about a third, and only by lowering the baseline.
The specific prompt's gap at 25 cruxes is 16.4 points, against 24.7 for a same-day rerun of the original prompt, a fall of 8.3 points (95\% interval 5.1 to 11.5, lower in 42 of 59 games).
The overlap stays at about 17\% under both prompts while the baseline falls from about 42\% to 33\%.
The remaining gap holds in 58 of 59 games.
Rewriting the multiple-rollout cruxes vaguer, or matching cruxes on the evidence behind them, leaves a gap (\cref{app:specific}).
The two sets therefore ask about different things, not merely at different levels of detail.

\subsection{Other Models and Roles}\label{sec:res_models}\label{sec:exp_models}

To check whether the gap is just a quirk of the GPT models, we also run the March Madness chain on three open-weight models, qwen3:32b, gemma4:31b and gpt-oss:120b.
Note that this means that different stages of the pipeline are being run by different models, which more faithfully reflects the way a deployed system would be built, as a combination of models for cost efficiency.

\paragraph{Other model families.} In March Madness the gap holds on the GPT runs and, in direction, on each of the three open-weight model families (\cref{tab:openweight}; \cref{app:openweight}).

\paragraph{Other roles.} All 27 cells of the grid of three predictors by three extractors at three levels fall below their baselines.
A model predicts the cruxes of its own forecasts at most marginally better than those of another model's forecasts (\cref{app:crossmodel}).

The gap survives every model family tested and every assignment of roles.

\section{Discussion}\label{sec:discussion}

In this paper we've shown that extracting the a posteriori cruxes of a model's forecasts is substantially different from predicting them in advance, and cheap enough to do for high-stakes forecasts, at an estimated 86 times less compute than a judge (\cref{app:costs}).
Hence, rather than relying on only qualitative definitions of what constitutes a ``crux'', there is instead an unsupervised method for extracting them from the model's own rollouts.
There are still many exciting questions to explore as we elaborate below, and we hope that this work will inspire future research in this area.

\paragraph{Can a model's cruxes be identified with finite compute?} Every crux counts equally here, so the overlap cannot tell a missed crux that would have moved the forecast from one that would not, and whether the gap survives weighting each crux by the value of information from resolving it is untested (\cref{app:designs}).
The gap also depends on which models produced it.
The predictor is a small model prompted a few ways, and in Metaculus it never saw the web results the rollouts drew on (\cref{sec:metaculus}).
The extractor changes the size of the gap (\cref{app:crossmodel}), and a single reranker decides every match, under instructions chosen by agreement with a language-model judge rather than with people (\cref{app:validation}).
That agreement was measured on March Madness alone, before the Metaculus source was added, so the cruxes counted here are the ones this extractor and this matcher agree on, and a costlier pipeline might find others.
Across 59 games and 15 events, mostly from GPT models, the direction of the result is firm and its size is not, since a second draw of the a priori cruxes moves the Metaculus gap by about ten points (\cref{app:alignment}).

\paragraph{Do a model's cruxes actually help a person decide?} A human study would be needed to see whether the a posteriori cruxes help a person make a better forecast.

\paragraph{What is a crux of an open-ended request?} Every request here is a forecast of one event, which gives two rollouts a number to disagree on.
However, reports, plans and code changes have no such number: is there a way to extract the cruxes of a model's reasoning for \emph{open-ended} requests?

\subsection*{AI use statement}

In this work, we used generative AI tools for execution of and giving feedback on the experiments, accessing and reformatting datasets, and visualization of results.
We have not used generative AI tools for proposing or refining hypotheses, designing the research methodology or experimental design, generating synthetic datasets, developing theoretical models or conceptual frameworks, formulating mathematical claims, implementing the critical software, or translation.
Proving mathematical claims, writing proofs, and qualitative or thematic data analysis are not applicable to this work.
Specifically, we used generative AI tools for creating and modifying figures, changing hyperparameters, drafting and editing explanatory prose of the paper, and recommending literature.
We have reviewed all AI-assisted work.
The critical software was written by the authors; every cited paper was checked for relevance and its citation sourced by an author; and every reported number, apart from the cost estimates marked as such, was recomputed from the recorded model outputs.
We take responsibility for the final content of this work, including text, claims or artifacts produced with the aid of generative AI.

\subsection*{Reproducibility statement}

\newcommand{\artifactlink}[1]{\ificlrfinal\url{#1}\else anonymous link\fi}
Two artifacts are public.
The extraction of cruxes on which forecasts take different sides, the one step that carries optimizations, is released as a Python package (\artifactlink{https://github.com/davidreber/forecast-cruxes}); the scouting and prediction stages are single prompts, given verbatim in \cref{app:prompts}, and they and the forecasting stage ship as scripts beside the data.
The data release (\artifactlink{https://github.com/davidreber/forecast-cruxes-reproduce}) holds every model output the paper analyzes; for Metaculus it holds question identifiers and a fetch script, since the site's terms do not permit redistributing its text.
The same repository reruns every stage downstream of the recorded model outputs from the released inputs alone, on independent hardware, and checks each reported number against a tolerance measured from drift between runs: raw reranker scores within 0.05, overlap and the gap below the baseline within 1.5 times the difference between two draws of the a priori cruxes.
The runs are dated (Metaculus forecasts 2026-02-28 to 03-01; March Madness scouting reports and forecasts 2026-03-17 to 03-18; multiple-rollout extractions 2026-03-19 and 03-31, the cross-model check's 2026-09-22 and the single-rollout cruxes' 2026-09-25; a priori cruxes 2026-03-31 for Metaculus, with a second draw on 2026-09-23, and 2026-09-21 and 09-22 for March Madness; the specific-prompt arms and their same-day controls 2026-09-24 for March Madness and 2026-09-24 to 09-25 for Metaculus), and no call set a temperature, seed or reasoning setting, so the API defaults applied.
The March Madness numbers on gpt-5.4-nano use a priori cruxes regenerated after an earlier version of the prediction prompt corrupted some of its cruxes; the Metaculus cruxes were not affected, and \cref{app:regeneration} shows that the corruption did not produce the finding.
\Cref{sec:forecasting} and \cref{tab:roles} specify the sources and models, \cref{sec:equivalence} the matching and clustering, \cref{sec:exp_headline} the overlap and baseline, \cref{app:robustness,app:costs,app:reports} the reranker instructions, costs and example scouting reports, and \cref{fig:prompts} quotes the load-bearing lines of each prompt.

\ificlrfinal
\subsubsection*{Acknowledgments}
This work is supported by ONR grant N00014-23-1-2591 and Open Philanthropy.
\fi

\bibliography{bibs/nlp_claim_decomposition,bibs/misc,bibs/llm_self_knowledge}
\bibliographystyle{bibs/iclr2027_conference}

\appendix

The appendices come in five parts: the robustness checks in detail, the models and costs, the data and prompts, some example model outputs, and related work in detail.

\section{Robustness Checks in Detail}\label{app:checks}

\subsection{The overlap, its baseline and the other scores}\label{app:metric}

Let the pooled cruxes fall into $J$ clusters of sizes $c_1, \dots, c_J$.
Call the a priori or single-rollout cruxes set A and the multiple-rollout cruxes set B, with $n_A$ and $n_B$ cruxes and $N = n_A + n_B$.
Let $a_j$ be the number of set A cruxes in cluster $j$.
The overlap is the share of clusters that hold cruxes of both sets,
\begin{equation}
\mathrm{overlap} = \frac{1}{J} \sum_{j=1}^{J} \mathbf{1}\big[\,0 < a_j < c_j\,\big].
\end{equation}
The baseline keeps the clusters fixed and reassigns the label A uniformly at random to $n_A$ of the $N$ pooled cruxes.
Cluster $j$ is then a random draw of $c_j$ of the $N$ cruxes, unshared only if the draw takes no set A crux or nothing but set A cruxes, so it is shared with probability
\begin{equation}
p_j = 1 - \frac{\binom{N - n_A}{c_j}}{\binom{N}{c_j}} - \frac{\binom{n_A}{c_j}}{\binom{N}{c_j}},
\qquad p_j = 0 \text{ when } c_j = 1,
\end{equation}
and the baseline is $\frac{1}{J}\sum_j p_j$, estimated as the mean overlap over 1{,}000 such shuffles.

The overlap and baseline coincide by construction at one crux per set, and the text reports sets of 5 to 25 cruxes.
A positive gap means the a priori cruxes land in the same clusters as the multiple-rollout cruxes less often than shuffled labels would, and a negative gap means more often.
Each game or event is paired with its own baseline, and a one-sided Wilcoxon signed-rank test over those pairs gives the $p$ value of each gap.

The single-rollout cruxes are scored the same way, with one adjustment: because they come from a single rollout, some multiple-rollout cruxes were extracted from pairs that include that rollout.
To account for that, the multiple-rollout cruxes that came from that rollout are removed before the overlap is scored.
The a priori cruxes are then rescored against the same reduced set of multiple-rollout cruxes, so the two are compared on equal terms (\cref{app:single}).

Two more scores use the same clusters and the same random relabelling.
Coverage is the share of the multiple-rollout cruxes whose cluster also holds an a priori or single-rollout crux.
Its mirror image is the share of the a priori or single-rollout cruxes whose cluster also holds a multiple-rollout crux.
A further check, the split-half check, splits one set, a priori or multiple-rollout, into two random halves and scores them against each other.
It answers the objection that the multiple-rollout cruxes are too varied for any set of cruxes to match (\cref{sec:res_headline}).

\subsection{Design detail for the checks}\label{app:designs}

\paragraph{What the gap below the baseline does not show.} The test does not show any a priori crux to be false (\cref{sec:discussion}).
They read as sensible questions about the game or event, and many may bear on its outcome.
An unmatched a priori crux may still split some pair of rollouts and fall among the multiple-rollout cruxes beyond the 25 scored.
The gap says nothing about any one crux.
The overlap counts every crux equally, so it cannot tell a missed crux that would change the forecast from one that would not.
If the cruxes absent from the a priori set would barely move the forecast, the gap would matter less to a person deciding than its size suggests.
Replacing the extractor gpt-5.4-nano with qwen3:32b on the same forecasts moves the gap at 25 cruxes by about 11 points (\cref{app:crossmodel}).
So the scoring can detect a change of that size.
One objection is that the reranker keeps apart a crux that names a factor and a crux that gives a number for the same fact.
The same replacement answers this objection in part.
From the same forecasts qwen3:32b writes 59 multiple-rollout cruxes per game against 160 for gpt-5.4-nano.
With qwen3:32b as the extractor, every combination of models and level still falls below its baseline.
The replacement changes the model as well as the number of cruxes, though.

\paragraph{Two open objections.} Two objections stay open.
One is weight, since every crux counts equally.
Weighting each crux by the value of information~\citep{howard1966information} from resolving it would answer this objection and is left to future work.
The other is vagueness, since part of the difference between the sets lies in how specifically the cruxes are written.
Asked for cruxes as specific as the multiple-rollout ones, the predictor complies, and the March Madness gap narrows from 24.7 to 16.4 points at 25 cruxes (\cref{sec:res_specificity,app:specific}).
The overlap stays at about 17\% and only the baseline falls, so the remaining 16 points are not a matter of wording.

\paragraph{Sources.} Rollouts that read different reports can split over a fact one report states and the other leaves out, so some March Madness cruxes may trace to how the reports were written.
A predictor that reads all the reports at once, rather than one on each team, could miss those cruxes, so a gap found only in March Madness could come from the reports.
In Metaculus the predictor sees less than the rollouts do (\cref{sec:metaculus}), so a gap found only there could come from what the predictor never saw.
The sources differ in more besides, including topic, how many rollouts each game or event gets, the prediction prompts, and which model plays each role (\cref{tab:roles}).
A difference in the gap's size could come from any feature the sources do not share.

\paragraph{Contested games.} Two rollouts can take opposite sides on a crux and still pick the same winner.
For example, one rollout may back a team because it expects the team to commit few turnovers, while another expects many turnovers and backs the same team for its rebounding.
A contested game also holds pairs that pick the same winner, so its multiple-rollout cruxes mix theirs with those of pairs that split on the winner.
The check shows whether the gap holds where rollouts do disagree over the winner, but it cannot isolate the cruxes of the pairs that split on it.

\paragraph{Strictness.} A strict reading splits cruxes that a loose reading would merge.
If the gap came from reading ``same crux'' too strictly, it could shrink under an instruction that asks for less.
Same question, same resolution and same dispute ask less of a match in turn.
A looser instruction is meant to score more pairs as matching, but each level's scores are clustered and cut anew, so a looser level need not end with larger clusters than a stricter one, and each level gets its own baseline (\cref{sec:exp_headline}).

\paragraph{Coarser clusters.} Same-resolution matching can keep a vague crux apart from its specific counterpart, merging two cruxes only when resolving one would necessarily resolve the other.
Learning how fast world demand for oil will grow settles whether it will still be growing, but learning only that it will still be growing leaves the rate open, so a vague crux and a specific one about the same fact can miss each other at that level.

In March Madness, at the same-resolution level, a sweep loosens matching without changing either set.
At each coarser cut two cruxes need to share less to fall together (\cref{app:alignment}).
Each cut gets its own baseline, computed on that cut's clusters.

\paragraph{Scope of the contested games.} In each contested game some rollouts pick different winners, so whether the predictor reads \emph{disagree} as taking different sides on a crux or as picking different winners, it names cruxes of a kind on which the rollouts do split.
A March Madness pick is a conclusion two rollouts either share or do not, and a Metaculus probability is not, so the check runs on March Madness only.

\paragraph{Choosing the instructions.} The three instructions, given in \cref{app:robustness}, were chosen from 10 candidates by agreement with a language model acting as a judge~\citep{zheng2023judging} (\cref{app:validation}).

\paragraph{Qualitative patterns.} In March Madness the quoted cruxes differ in which part of the game they concern as well as in how precisely they state it.
For Iowa State against Kentucky, the a priori crux quoted in \cref{fig:composite_venns} asks whether free-throw performance becomes a decisive factor.
The multiple-rollout cruxes include some whose assumptions state the number of steals by Tamin Lipsey, Iowa State's point guard, and the number of turnovers Iowa State forced in an earlier game.
These three quoted cruxes alone cannot show which of the two differences keeps them from matching.

On Metaculus, multiple-rollout cruxes can cite facts that appear nowhere in the prediction prompt.
For the Frontier event of \cref{fig:composite_venns}, both sets name cash on hand, debts coming due and demand for air travel.
The a priori cruxes stay general, asking whether Frontier keeps enough cash to avoid filing for bankruptcy and whether weak demand for air travel hurts its revenue.
The multiple-rollout cruxes cite the bankruptcy restructuring of Spirit Airlines, another budget airline in the United States, and Frontier's Altman Z-score.
The rollouts found both through web search, which the predictor does not see (\cref{sec:metaculus}).
The predictor could have named them only from what it learned in training.

\paragraph{Specific prompts.} A crux can name a factor, such as turnovers, make a claim about it, such as that Kentucky turns the ball over often, or state a value, such as Kentucky's turnover rate of 12.0\%.
A local model tags every crux in the deduplicated pools with one of these three levels, and the tags show where the two sets differ.
In March Madness 66\% of a priori cruxes make a claim and 32\% state a value, while 91\% of multiple-rollout cruxes state a value.
In Metaculus 89\% of a priori cruxes make a claim and 44\% of multiple-rollout cruxes state a value.
Almost no crux on either side names a bare factor.

\paragraph{Other models.} The open-weight and mixed runs use other extractors.
The extractor changes the size of a gap, so these runs are compared with the main result in direction only.

\subsection{Robustness across levels of strictness}\label{app:robustness}
\suppressfloats[t]

\begin{figure}[tp]
\centering
\includegraphics[width=\linewidth]{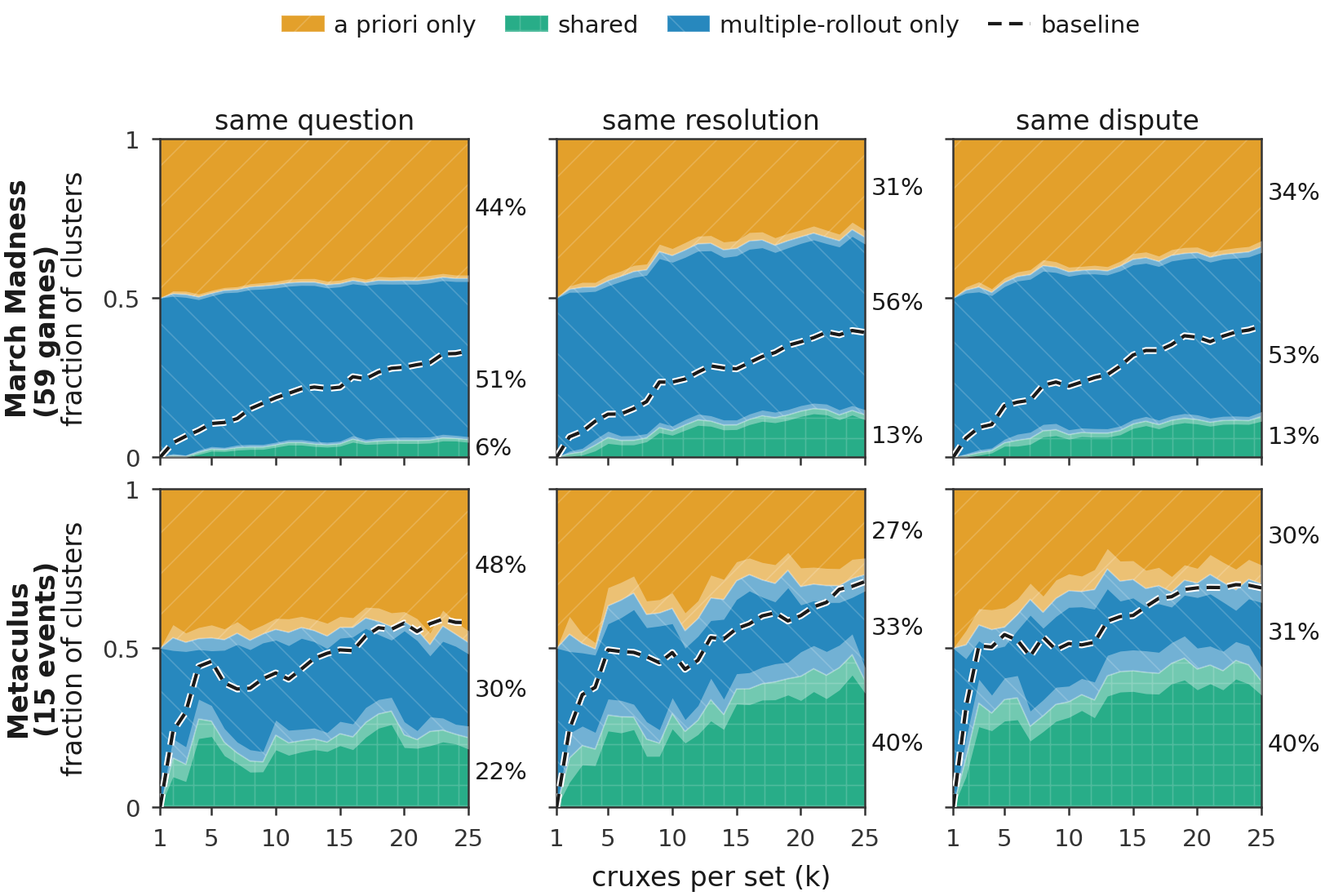}
\caption{Whether two cruxes must ask the same question or need only concern the same dispute, predicted cruxes match discovered ones less often than the permutation baseline.
Stacked-area decomposition at the three levels, same question, same resolution and same dispute, for March Madness (59 games, top row) and Metaculus (15 events, bottom row), in the colours of \cref{fig:stacked}.
The numbers at the right of each panel give its three shares at 25 cruxes.
At 25 cruxes the March Madness overlap is 5.6\%, 13.4\% and 12.8\% against baselines of 33.4\%, 39.2\% and 41.2\%, and the Metaculus overlap is 21.9\%, 39.9\% and 39.7\% against 58.0\%, 70.9\% and 68.9\%.}
\label{fig:stacked_p0}\label{fig:stacked_p9}
\end{figure}

The gap below the baseline holds at all three levels of strictness, though its size changes with the level.
\Cref{fig:stacked_p0} shows the decomposition at all three levels in both sources, and its middle column repeats \cref{fig:stacked}.
In March Madness the shared band stays the thinnest of the three at every level.
\Cref{tab:matching_criteria} compares the three levels in both sources at $k = 25$.

\paragraph{An example of the three levels.}
The oil cruxes show what the broad level adds.
Whether world demand for oil will still be growing in ten years and whether it will have peaked by then ask one question, so they match at every level.
Whether electric cars will make up most new car sales in ten years can be settled without settling demand, so it misses the demand cruxes at the same-resolution level.
Two rollouts that disagree over demand because one expects electric cars to take over could have that dispute ended by settling either, so at the broad level it matches them.

\paragraph{Level and cut.}
The level changes what the reranker is asked and the cut changes how many merges are kept, so the two loosen matching in different ways, and the main results fix both.

\paragraph{Which cruxes are scored.}
Each set of cruxes is shuffled once, with a seed derived from the game or event and the set, and the first $k$ cruxes of that shuffle enter, so the cruxes scored at 12 are among those scored at 25.
A crux that appears more than once in a set is kept once before the set is scored, and \cref{app:openweight} names the runs that skip this step.
Every set runs past 25 cruxes: after repeats are removed, March Madness sets hold 63 to 86 a priori and 137 to 197 multiple-rollout cruxes per game, and Metaculus sets 243 to 359 and 907 to 1{,}714 per event, so only part of each set is scored, and at 25 a Metaculus set contributes a small sample of what it holds.

\begin{table}[tp]
\caption{At every level of strictness the gap below the permutation baseline is significant in both sources, and nearly every game and event falls below its own baseline.
Overlap and baseline at $k = 25$ at each level of strictness, each level against its own baseline, for March Madness (59 games) and Metaculus (15 events).
Overlaps carry $\pm$ 1 SE across games or events.
Gaps are in points, and $p$ is from a one-sided Wilcoxon signed-rank test across games or events.
The last column counts games or events below their own baseline; the rest equal it, apart from one March Madness game at the same-resolution level that lies above.}
\label{tab:matching_criteria}
\begin{center}
\small
\small\begin{tabular}{@{}llccccc@{}}
\toprule
Source & Level & Overlap (\%) & Baseline (\%) & Gap & $p$ & Below \\
\midrule
March Madness & same question & $5.6 \pm 0.9$ & 33.4 & 27.8 & $< 10^{-10}$ & 59 / 59 \\
 & same resolution & $13.4 \pm 1.6$ & 39.2 & 25.9 & $< 10^{-10}$ & 55 / 59 \\
 & same dispute & $12.8 \pm 1.5$ & 41.2 & 28.4 & $< 10^{-10}$ & 57 / 59 \\
\midrule
Metaculus & same question & $21.9 \pm 3.6$ & 58.0 & 36.1 & $6.1 \times 10^{-5}$ & 14 / 15 \\
 & same resolution & $39.9 \pm 4.0$ & 70.9 & 31.0 & $1.2 \times 10^{-4}$ & 13 / 15 \\
 & same dispute & $39.7 \pm 4.4$ & 68.9 & 29.3 & $2.4 \times 10^{-4}$ & 12 / 15 \\
\bottomrule
\end{tabular}
\end{center}
\end{table}

The three boxes below give the reranker instructions for each level, in order of strictness.
Each box prints the level's positive instruction, which asks whether two cruxes match, and its negative instruction, which asks whether they differ.
Every reported number uses the positive instruction only, scored in both orders as \cref{sec:equivalence} describes.
A multiple-rollout crux records which of its two answers, A or B, each of its two rollouts assumed.
The predictor writes an a priori crux without seeing any rollout, so its labels follow no rollout.
An a priori crux can put first the answer that a matching multiple-rollout crux puts second.
Every positive instruction therefore tells the reranker that the labels may be swapped.

\begin{CodeBox}[Reranker: same-question matching]
\textbf{Positive instruction.} Given a crux question about a forecasting problem, determine whether the following crux question addresses the same underlying point of disagreement, even if worded differently. The answer labels (A/B) may be swapped.

\textbf{Negative instruction.} Given a crux question about a forecasting problem, determine whether the following crux question addresses a DIFFERENT point of disagreement. Consider them different even if they share vocabulary, as long as the core factual question is distinct.
\end{CodeBox}

\begin{CodeBox}[Reranker: same-resolution matching]
\textbf{Positive instruction.} Two crux questions are equivalent if resolving one would necessarily resolve the other, because they ask about the same underlying factual question. Ignore differences in wording, specificity, or answer-label ordering (A/B may be swapped). Does the following crux question target the same factual uncertainty?

\textbf{Negative instruction.} Two crux questions are different if they could be resolved independently --- knowing the answer to one would not determine the answer to the other. Even if they use similar vocabulary or concern the same topic, determine whether the following crux question targets a genuinely different factual uncertainty. Answer labels (A/B) may be swapped.
\end{CodeBox}

\begin{CodeBox}[Reranker: same-dispute matching]
\textbf{Positive instruction.} Imagine two forecasters disagreeing. If resolving the query crux would settle their specific point of contention, would resolving the following crux question settle the exact same point of contention? Two cruxes are equivalent if they would end the same debate. Ignore wording differences and answer-label order (A/B may be swapped).

\textbf{Negative instruction.} Imagine two forecasters disagreeing. If resolving the query crux would settle one point of contention, would the following crux question address a different point of contention --- one that could persist even after the first crux is resolved? Two cruxes are different if they would end different debates. Answer labels (A/B) may be swapped.
\end{CodeBox}

\paragraph{Overlap at the strict and broad levels.}
\Cref{fig:stacked_p0} shows the overlap at all three levels as each set grows.
In both sources the broad level gives a slightly lower overlap than the same-resolution level.

\paragraph{Validation.}\label{app:validation}
The three reranker instructions above were chosen from 10 candidate instructions by balanced accuracy on 2{,}000 crux pairs that gpt-5.4-nano, as a judge with a two-step rubric (map the two cruxes' answers onto each other, then compare their scope), had labeled as matching or not.
The pairs were drawn at random from the pooled 25 plus 25 cruxes of the 45 resolved March Madness games of an earlier run, 1{,}018 of them across the two sets and 982 within one set, and the judge called 34 matching.
Balanced accuracy is the mean of two recalls, the share of judge-matched pairs the cut puts in one cluster and the share of judge-distinct pairs it keeps apart.
Same question scored 0.845, same resolution 0.826 and same dispute 0.840.
The two recalls behind those means are 0.706 and 0.984 for same question, 0.676 and 0.976 for same resolution, and 0.735 and 0.945 for same dispute, so the cut keeps 94 to 98 percent of judge-distinct pairs apart.
Only 3 of the 34 pairs the judge marked as equivalent cross the two sets, so the judge's accuracy on cross-set pairs, the ones that decide the overlap, cannot be measured from this set.

The judge's labels have not been compared directly with human labels, so the instructions were chosen by agreement with a language model and not with human judgment.

\subsection{Coarser cuts and split halves}\label{app:alignment}

Two objections to the headline number can be answered from the same scored cruxes, without new model calls.
The first is that the low overlap may be an artifact of where the clustering is cut, so that looser matching would let the two sets share more clusters (\cref{sec:res_specificity}).
The second is that the multiple-rollout cruxes may be too varied for any set of cruxes to match them (\cref{sec:res_headline}).

\begin{figure}[t]
\centering
\includegraphics[width=0.8\linewidth]{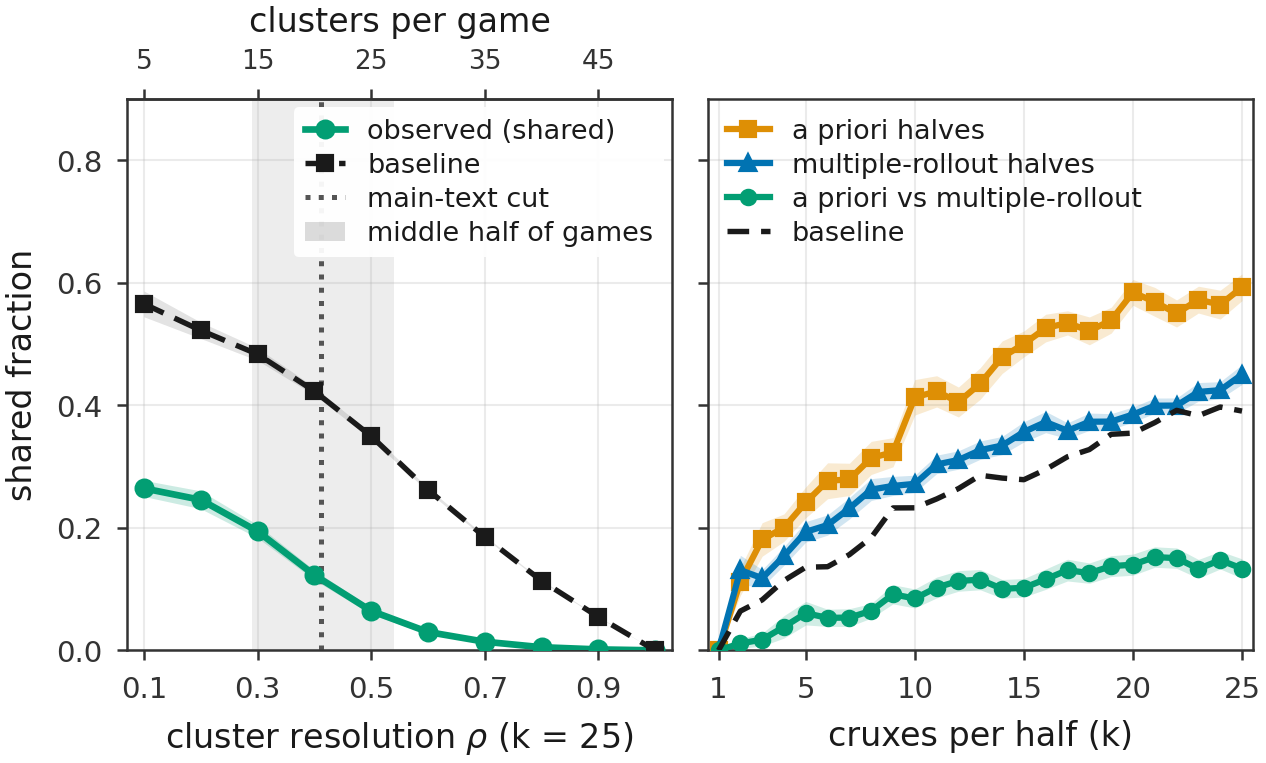}
\caption{In March Madness, however coarsely cruxes are grouped, predicted cruxes match discovered ones less often than the permutation baseline, though halves of the discovered cruxes match each other.
\textbf{Left:} Coarsening the clustering does not close the gap below the baseline.
The overlap and its baseline across cuts of the tree, from broad categories on the left to one crux per cluster on the right (March Madness, same-resolution matching, 25 cruxes per set).
The horizontal axis is $\rho$, the number of clusters divided by the number of pooled cruxes, and the top axis gives the clusters per game.
The dotted line marks the cut used in the main text, a mean of 20.5 clusters per game ($\rho = 0.41$), and the grey band the middle half of games.
\textbf{Right:} Two halves of either set share several times the overlap the two sets share with each other.
Overlap against the number of cruxes drawn from each set, for two halves of the a priori cruxes, two halves of the multiple-rollout cruxes, and the a priori cruxes against the multiple-rollout cruxes.
The dashed line gives the permutation baseline for the a priori against multiple-rollout comparison (March Madness, same-resolution matching).
At 25 cruxes two halves of the multiple-rollout cruxes sit just above the dashed baseline, two halves of the a priori cruxes well above it, and the two sets far below it.
So the discovered cruxes are not too scattered for a set of cruxes to match them.}
\label{fig:granularity}\label{fig:withinset}
\end{figure}

\paragraph{Overlap at every cut.} Cutting one tree at a series of points changes how loosely cruxes are matched.
The pooled a priori and multiple-rollout cruxes are clustered once.
That tree is then cut at cluster counts running from a small fraction of the cruxes up to one cluster per crux.
A coarse cut lumps cruxes into broad categories.
The finest cut leaves every crux alone.
Only the cut moves, so re-clustering cannot confound the sweep.
\Cref{fig:granularity} draws the sweep, and \cref{sec:res_specificity} reports what it shows.

\paragraph{Two halves of one set.} Each set is clustered on its own, split at random into two halves, and the two halves are compared exactly as the a priori and multiple-rollout cruxes are.
Averaged over random splits, the split-half overlap is the permutation baseline applied to one set alone rather than to the pooled pair.
\Cref{fig:withinset} draws the check, and \cref{sec:res_headline} reports what it shows.
Two halves of the a priori cruxes overlap 4.4 times as much, so the predictor's own cruxes also cluster with one another.

\begin{figure}[tp]
\centering
\includegraphics[width=\linewidth]{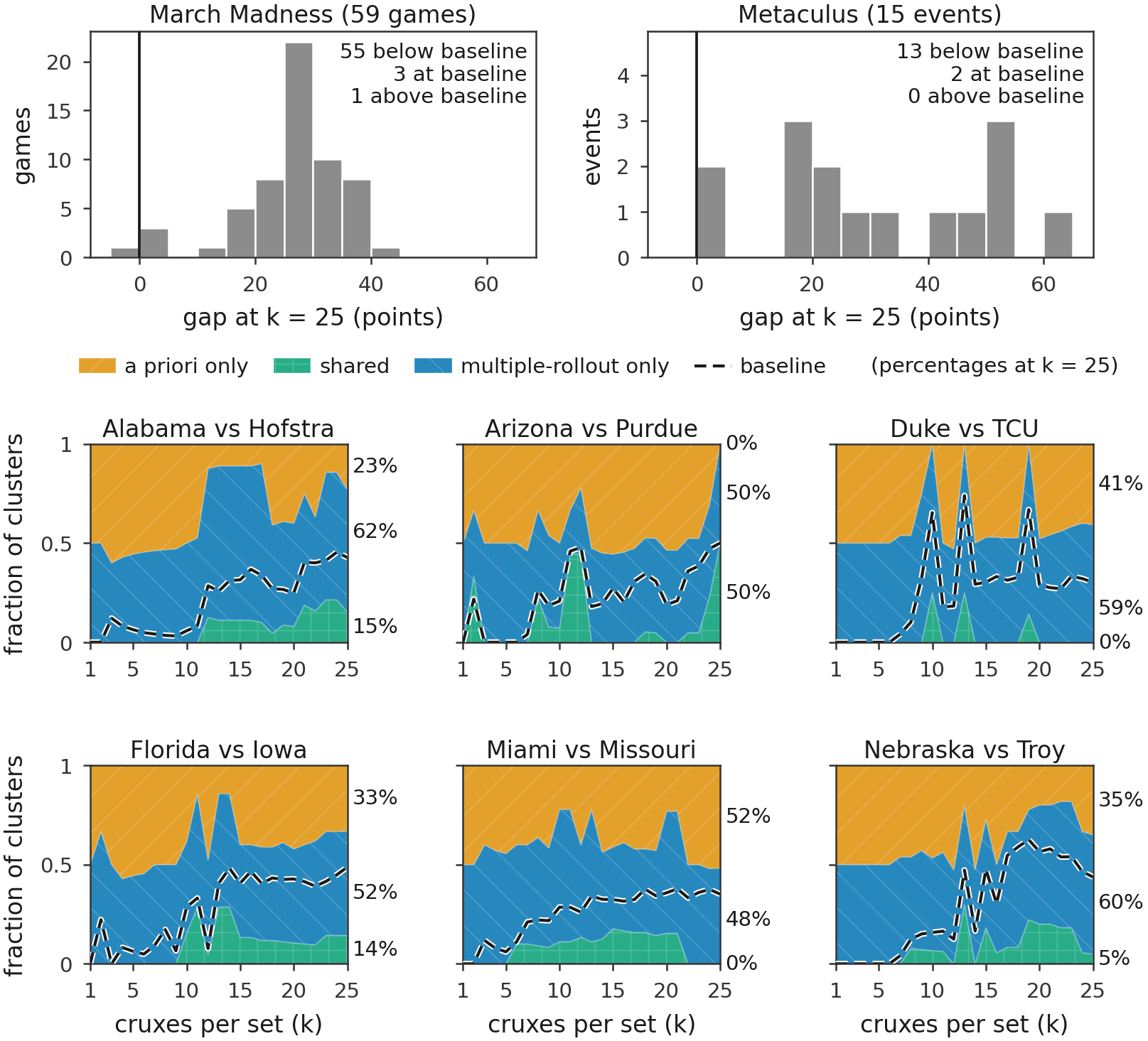}
\caption{Predicted cruxes match discovered ones less often than the permutation baseline in nearly every game and event, so the mismatch is the typical case.
\textbf{Top:} the gap of each game and event at 25 cruxes, its own baseline minus its overlap, with a line at zero.
In March Madness 55 of 59 games fall below their own baseline, 3 equal it and 1 lies above it; in Metaculus 13 of 15 events fall below, 2 equal it and none lies above.
\textbf{Bottom:} six March Madness games chosen at random (seed 42), as each set grows to 25 cruxes, in the colours of \cref{fig:stacked} and with each game's own baseline as the dashed line.
The shared band rises and falls with set size and differs from game to game.
Same-resolution matching throughout.}
\label{fig:per_matchup_hist}
\end{figure}

\paragraph{Baselines and single games in March Madness.}
The baseline rises because a larger pool gives clusters more cruxes, and a cluster with more cruxes is more likely to receive both labels when they are shuffled.
Here below means lower than the mean of the game's own shuffles, a comparison of two numbers and not a test run on each game.
A game can equal its baseline when shuffling the labels cannot change its overlap; the three games that do so each end with two clusters, one holding every crux but one, so no shuffle moves the overlap.
The median game shares 10.5\% of its clusters and nine share none, so the mean of 13.4\% overstates what a typical game shares.

\paragraph{Metaculus events.}
Metaculus rollouts resemble what a person gets by running one forecasting system on one request many times.
In Metaculus the predictor is not the forecaster, so the gap holds when a model predicts another model's cruxes, while the test of a model predicting its own cruxes rests on March Madness.

In Metaculus, as in March Madness, the gap holds for sets of 5 to 25 cruxes (\cref{fig:stacked}, right).
It is already there at 5 cruxes, where the overlap is 29.0\% $\pm$ 5.0\% (SE across 15 events) against a baseline of 49.5\% ($p = 0.0015$, one-sided Wilcoxon signed-rank test).
Metaculus baselines run higher than those of March Madness, so the two sources are compared by their gaps.
In Metaculus, 13 of the 15 events fall below their own baseline.
The other two equal theirs, since the cut leaves each with exactly two clusters, 49 cruxes in one and a lone multiple-rollout crux in the other, so no shuffle of the labels can move the overlap.
The gap averaged over events therefore describes the typical event.

The $\pm$ 4.9 on the Metaculus gap of 31.0 points (\cref{sec:res_variation}) covers variation across events only.
A second draw of the a priori cruxes moves the gap by about ten points (10.6 at the same-resolution level and 25 cruxes), so with only 15 events its size is unstable.
Both draws fall below the baseline at every strictness level and for sets of 5 to 25 cruxes, so its direction is stable.

The overlap varies widely across events, with a standard deviation of 15.6 points.

\paragraph{Set size and spread.}
In March Madness the overlap rises from 6.3\% $\pm$ 1.9\% (SE) at 5 cruxes, against a baseline of 13.5\%, to 13.4\% at 25, against 39.2\%.
The overlap varies widely across games, with a standard deviation of 12.1 points (\cref{fig:per_matchup_hist}, bottom).
Iowa State vs.\ Kentucky, the game in the left panel of \cref{fig:composite_venns}, shares 7.1\%.
The Frontier Group event, in the right panel of \cref{fig:composite_venns}, lies at the low end, with an overlap of 8.3\%.

\subsection{Specific prompts and coarsening}\label{app:specific}

\Cref{sec:exp_specific} names the check, and this appendix gives the detail and numbers of its arms, all at the same-resolution level.
Each crux in the deduplicated pools was tagged as naming a factor, making a claim or stating a value by qwen3:32b run locally.

The predictor is re-run with a prompt that asks it to state each crux at the specificity of a value, and its cruxes are scored exactly as in \cref{sec:exp_headline}.
A control arm re-runs the original prompt on the same day, so that drift between draws can be told from the effect of the prompt.
Two further arms are reported for completeness: an exemplar arm that shows the predictor cruxes extracted from another game's or event's rollouts, which changes the writing style as well as the specificity, and a ladder of three prompts asking for factor, claim and value cruxes in turn.
The same tags check whether each prompt did what it asked.

The coarsening check works from the other side, on the cruxes as written.
A local model rewrites each multiple-rollout crux one level vaguer, mostly from a value to a claim, with the a priori cruxes untouched, and in a second arm rewrites both sets toward the factor level.
A third comparison restricts both sets to cruxes tagged at the same level and scores the restricted pools on the games where both sets hold enough.

\Cref{tab:specific_tags} gives the shares.

\begin{table}[t]
\caption{Asked for specific cruxes, the predictor states values at least as often as the discovered cruxes do, while its usual cruxes mostly do not.
Share of cruxes at each level of specificity, in percent, for the cruxes as written and for each prompt.
The factor prompt of the ladder failed its check, so the ladder separates claims from values only.}
\label{tab:specific_tags}
\begin{center}
\small
\small\begin{tabular}{@{}llccc@{}}
\toprule
Source & Cruxes & Factor & Claim & Value \\
\midrule
March Madness & predicted, original prompt & 1.9 & 66.4 & 31.7 \\
 & predicted, same-day control & 1.8 & 66.5 & 31.7 \\
 & predicted, specific prompt & 0.1 & 3.3 & 96.6 \\
 & predicted, exemplar prompt & 0.2 & 2.8 & 97.0 \\
 & predicted, ladder, factor prompt & 9.3 & 68.1 & 22.6 \\
 & predicted, ladder, claim prompt & 1.9 & 77.2 & 20.9 \\
 & predicted, ladder, value prompt & 0.3 & 4.7 & 95.0 \\
 & extracted (multiple-rollout) & 0.1 & 9.3 & 90.6 \\
Metaculus & predicted, original prompt & 0.6 & 88.8 & 10.6 \\
 & predicted, same-day control & 0.4 & 87.2 & 12.4 \\
 & predicted, specific prompt & 0.1 & 14.2 & 85.7 \\
 & predicted, exemplar prompt & 0.1 & 18.2 & 81.6 \\
 & extracted (multiple-rollout) & 0.8 & 55.6 & 43.6 \\
\bottomrule
\end{tabular}
\end{center}
\end{table}

\begin{table}[t]
\caption{In March Madness, asking for specific cruxes narrows the gap by about a third only by lowering the permutation baseline, and no prompt closes it.
Gap in points under each prompt, at 25 and 12 cruxes per set.
March Madness $p$ values are two-sided Wilcoxon signed-rank tests of control minus arm over 59 games.
The $p$ values are normal approximations.
In Metaculus the drift between the original and the same-day control (7.8 points at $k = 25$) exceeds every difference between arms.
Every row, the original included, uses the seed of the arm runs for the baseline's label shuffles.
That seed is the same for every game or event.
The runs of \cref{tab:matching_criteria} instead give each game or event its own seed.
With that seeding the original prompt's gaps are 25.9 and 15.0 in March Madness and 31.0 and 18.8 in Metaculus.}
\label{tab:specific_arms}
\begin{center}
\small
\small\begin{tabular}{@{}llcc@{}}
\toprule
Source & Prompt & $k = 25$ & $k = 12$ \\
\midrule
March Madness & original (headline, one seed) & 25.8 & 14.8 \\
 & same-day control & 24.7 & 15.0 \\
 & specific & 16.4 ($p = 1.4 \times 10^{-5}$) & 10.1 ($p = 0.0012$) \\
 & exemplar & 12.1 ($p = 1.4 \times 10^{-8}$) & 7.8 ($p = 5.2 \times 10^{-6}$) \\
 & ladder, factor & 27.4 ($p = 0.10$) & 16.1 ($p = 0.57$) \\
 & ladder, claim & 23.0 ($p = 0.40$) & 12.5 ($p = 0.043$) \\
 & ladder, value & 15.7 ($p = 2.7 \times 10^{-6}$) & 8.4 ($p = 3.2 \times 10^{-5}$) \\
Metaculus & original (headline, one seed) & 31.4 & 18.7 \\
 & same-day control & 23.6 & 24.1 \\
 & specific & 31.0 & 16.9 \\
 & exemplar & 23.3 & 17.2 \\
\bottomrule
\end{tabular}
\end{center}
\end{table}

Under the specific prompt the March Madness overlap is 0.169 in both the control and the specific arm, and the baseline falls from 0.417 to 0.333.
The exemplar arm sits 4.3 points below the specific arm ($p = 3 \times 10^{-4}$), but the writing style of the extractor and the facts of the donor game enter the cruxes it predicts, so the specific arm is the clean number.
The same-day control's overlap moved 0.036 from the headline draw, a hair outside the 0.0355 tolerance the run uses for drift, while its gap moved 1.1 points.
In Metaculus, 2\% of control cruxes and 6\% of specific and exemplar cruxes came from calls whose question and answer arrays had unequal lengths; scoring without them gives the same reading.

\Cref{tab:specific_arms} gives the specific-prompt arms.
\Cref{tab:coarsening} gives the coarsening arms.
The rewrites toward the factor level mostly stopped at the claim level (in an automated spot check, 9 of 9 value cruxes reached a claim, 3 of 11 reached a factor, and 0 of 10 claims reached a factor), so the both-sides arm tests claim-level coarsening of both sets, and one local model rewrote both sides, so a shared writing style can inflate its overlap.
No arm moves the gap toward zero; the smallest gap in any arm is 11.0 points against its own baseline ($p = 2.3 \times 10^{-9}$, exact one-sided Wilcoxon signed-rank test).

\begin{table}[t]
\caption{Making cruxes vaguer, or comparing only cruxes of equal specificity, leaves predicted cruxes matching discovered ones less often than the permutation baseline.
Gap in points when cruxes are rewritten to be vaguer, and when both sets keep only cruxes at the same levels of specificity.
The $p$ values are two-sided Wilcoxon signed-rank tests of arm against original over games or events.
The original column uses the full set of multiple-rollout cruxes, none rewritten or removed, on the same games or events.
Both columns shuffle the two labels with one fixed seed for all games or events.
So the original column gives 25.8 and 14.8 in March Madness and 31.4 in Metaculus.
The main text gives each game or event its own seed and reports 25.9, 15.0 and 31.0.}
\label{tab:coarsening}
\begin{center}
\small
\setlength{\tabcolsep}{4pt}
\small\begin{tabular}{@{}llccc@{}}
\toprule
Source & Arm & $k$ & Original & Arm \\
\midrule
March Madness & multiple-rollout cruxes, rewritten one level vaguer & 25 & 25.8 & 26.9 ($p = 0.57$) \\
 & & 12 & 14.8 & 16.8 \\
 & both sets rewritten toward factors & 25 & 25.8 & 15.3 ($p = 1.3 \times 10^{-6}$) \\
 & & 12 & 14.8 & 11.0 ($p = 0.008$) \\
 & both sets restricted to values (24 games) & 25 & 23.9 & 22.4 \\
 & both sets restricted to values (53 games) & 12 & 15.6 & 15.1 \\
 & both sets restricted to claims or values & 25 & 25.8 & 28.4 \\
Metaculus & multiple-rollout cruxes, rewritten one level vaguer & 25 & 31.4 & 25.2 \\
 & both sets rewritten toward factors & 25 & 31.4 & 29.4 \\
 & both sets restricted to values (7 events) & 25 & 24.3 & 22.9 \\
\bottomrule
\end{tabular}
\end{center}
\end{table}

\paragraph{Coarser cuts and specific prompts.}
Overlap and baseline meet only at one crux per cluster, where no cluster can hold cruxes from both sets.
Looser matching therefore does not close the gap.

\paragraph{Evidence matching.} A third granularity check asks whether the gap shrinks when cruxes are matched on the evidence they rest on rather than on their wording.
In two batch runs, gpt-5.4-nano wrote a statement of the evidence behind every crux in both sets.
The overlap was then computed again, with the reranker matching these statements instead of the cruxes' own wording.
The reranker used one wording of its instruction per level, each chosen by agreement with a language-model judge, qwen3:32b.
As an anchor, the reranker also matched the same cruxes on their own wording, under the same-resolution instruction of the main text.
The clustering, the cut, the seeding and the baseline are computed as in the main results.
The judge and the reranker both ran on a local machine.
\Cref{tab:evidence} gives the gaps for the 59 March Madness games.
Five of the six differences from the anchor lie inside the run's drift tolerance of 3.55 points; the exception is the moderate wording at 12 cruxes, where the gap is 4.4 points larger (95\% interval 1.6 to 7.2, exact $p = 0.005$).
Between 45 and 58 of the 59 games stay below the baseline under every wording (exact one-sided $p$ at most $3 \times 10^{-14}$), so matching on evidence does not close the gap, and where it moves it, it widens it.
All six wordings were scored in full, and the three reported were then chosen on 580 judged pairs; only the strict pick is settled by the data (balanced accuracy 0.906 against 0.804 for its alternative), while the moderate and broad picks lie within their intervals of their alternatives (0.664 against 0.645, 0.642 against 0.603), and the alternative wordings give gaps of 22.5, 19.2 and 16.4 points at 25 cruxes and 16.9, 19.7 and 22.7 at 12, so the reading does not depend on the pick.
Two diagnostics bear on the reading.
The judge finds almost no evidence matches across the two sets: of 300 cross-set pairs it called 1 a strict match, 11 moderate and 22 broad, so the calibration sample had to be filled with within-set pairs (280 pairs), and of the 2,363 cross-set pairs the paper's instruction matches on wording, 11.9\%, 19.0\% and 43.1\% also match on their written evidence at the three wordings.
A keyword proxy says why: 75\% of the multiple-rollout cruxes' evidence statements name a season statistic, rank or record, against 11\% of the a priori cruxes'.
The two sets rest their cruxes on different kinds of evidence, not only on different wording.

\begin{table}[t]
\caption{Matching cruxes on the evidence behind them, instead of on their wording, still leaves predicted cruxes matching discovered ones less often than the permutation baseline.
Gap in points when cruxes are matched on their written evidence under three wordings of the matching instruction, beside matching on the cruxes' own text on the same pools (March Madness, 59 games).
The full set of multiple-rollout cruxes gives 25.9 points at 25 cruxes.}
\label{tab:evidence}
\centering
\small
\begin{tabular}{@{}lrrrr@{}}
\toprule
Cruxes per set & Strict & Moderate & Broad & Text, same pools \\
\midrule
25 & 20.3 & 24.3 & 20.1 & 22.9 \\
12 & 13.9 & 18.2 & 16.0 & 13.7 \\
\bottomrule
\end{tabular}
\end{table}

The specific prompt does what it asks, since 97\% of the cruxes it returns state a value, against 32\% under the original prompt.
The specific prompt leaves a gap of 16.4 points and the ladder's value prompt one of 15.7.
The exemplar prompt narrows the gap further, to 12.1 points, though it changes the writing style along with the specificity.
In Metaculus every arm stays below its baseline (the specific prompt in 13 of 15 events), and the differences between arms sit inside the drift between two draws, so 15 events can neither confirm nor contradict the March Madness fall.

Rewriting each multiple-rollout crux one level vaguer, with the a priori cruxes untouched, gives 26.9 points against 25.8; restricting both sets to cruxes that state values gives 22.4 against 23.9 on the 24 games where both sets hold enough.
Only rewriting both sets with the same model narrows the gap, to 15.3 points, still far below the baseline, and that arm also gives both sets one writing style.

\subsection{The single-rollout cruxes against the multiple-rollout cruxes}\label{app:single}

\paragraph{Selection and leakage.} One stored forecast per game or event is picked by a seeded rule (seed 42 hashed with the question identifier, modulo the number of forecasts) and the single-rollout cruxes are extracted from it on their own with a single-forecast prompt (\cref{app:prompts}).
The index counts into the forecast files in sorted file-name order, not numeric order, so index 0 is \texttt{forecast.json} and index 1 is \texttt{forecast10.json}.
The extractor is the same pinned model as the multiple-rollout extraction in each source, gpt-5.4-nano-2026-03-17 for March Madness and gpt-5.4-mini-2026-03-17 for Metaculus, through the batch API with a 5,000-token output cap and no sampling parameter set, and every stored forecast was rendered exactly as the pairwise extractor saw it.
The prompt carries the pairwise prompt's rubric and returns a strict JSON object with an analysis of the forecast and at most 30 cruxes, most important first, each a neutral question with the premise the forecast relies on and its contrary; the single-rollout cruxes are the questions in the returned order.
A forecast yields about 12 cruxes in March Madness and about 6 in Metaculus, always fewer than 25, so at $k = 25$ the whole forecast is scored against 25 pooled cruxes and the baseline is computed with that smaller label count.
In the headline pool, 22.7\% of the 25 multiple-rollout cruxes scored per March Madness game (5.3\% per Metaculus event) came from pairs that include the chosen forecast, and removing them gives the stripped multiple-rollout cruxes.
They are removed, and both the single-rollout cruxes and the a priori cruxes are scored against those stripped multiple-rollout cruxes, since removing items changes which 25 pooled cruxes the seeded selection draws.
That alone moves the gap of the Metaculus a priori cruxes from 31.0 points on the full multiple-rollout cruxes to 21.9 on the stripped multiple-rollout cruxes.
In March Madness the leakage mattered for the single-rollout cruxes (gap 3.2 points with the leaked items in, 6.4 with them out); in Metaculus the conclusion is the same either way.
Everything else, the reranker, the instructions, the clustering, the cut, the deduplication and the shuffle, is identical to the headline run.

\Cref{tab:prompt_diff} lists what changes between the pairwise extraction prompt and the single-forecast prompt of \cref{app:prompts}; the four-criterion rubric, the world-fact and neutrality rules, the output shape and the closing quality reminder are the same.

\begin{table}[H]
\caption{The single-forecast extraction prompt asks what one forecast rests on where the pairwise prompt asks where two forecasts disagree.
Every part of the two prompts that differs, in descending order of how much the change matters; all else is shared.}
\label{tab:prompt_diff}
\begin{center}
\small
\begin{tabular}{@{}p{2.4cm}p{5.4cm}p{5.4cm}@{}}
\toprule
Part of the prompt & Pairwise (multiple-rollout) extraction & Single-forecast (single-rollout) extraction \\
\midrule
Task statement & none; the rubric leads straight into the steps & ``You are given ONE argument: a single forecast for the question below, with its reasoning. Identify the cruxes of this argument: the factual premises about the world that its forecast turns on, so that if a premise were false, the forecast would change.'' \\
Distillation & ``Distill any differences into cruxes'' & ``Distill the premises its forecast depends on into cruxes'' \\
\raggedright option\_A and option\_B & Argument A's premise and Argument B's premise, contradicting along the crux dimension & the premise this argument relies on and the contrary premise, ``so that if it were true the forecast would change'' \\
Analysis step & analyze both arguments, do the same analysis for Argument B, then for each dimension note whether the arguments differ & analyze the one argument on the same four dimensions, with no comparison step \\
\raggedright Maximum and order & none stated & ``Return at most 30 cruxes, most important first: the premises whose reversal would move this forecast the most'' \\
Minimum output & ``You MUST return at least one crux if the arguments rely on different world-facts, even if they reach the same final answer'' & no minimum \\
Per-criterion rationale & required for each crux & dropped \\
\bottomrule
\end{tabular}
\end{center}
\end{table}

\begin{table}[t]
\caption{Cruxes from one forecast come far closer to the permutation baseline than predicted cruxes, so scoring one set of cruxes against a pooled one does not explain the gap.
Overlap against baseline at $k = 25$, same resolution, the single-rollout cruxes and the a priori cruxes scored on the same stripped multiple-rollout cruxes.
Gaps are in points, the counts are games or events below, above and at their own baseline, and $p$ is from the one-sided exact Wilcoxon test.}
\label{tab:single}
\centering
\small
\begin{tabular}{@{}lllll@{}}
\toprule
Source & Cruxes & Overlap vs baseline & Gap & Below / above / tied, $p$ \\
\midrule
March Madness & single-rollout & 18.8 vs 25.1 & 6.4 & 47 / 11 / 1, $4 \times 10^{-9}$ \\
March Madness & a priori & 14.6 vs 41.4 & 26.8 & 56 / 0 / 3, $10^{-17}$ \\
Metaculus & single-rollout & 48.5 vs 44.0 & $-4.5$ & 2 / 9 / 4, 0.94 \\
Metaculus & a priori & 35.1 vs 56.9 & 21.9 & 11 / 1 / 3, $5 \times 10^{-4}$ \\
\bottomrule
\end{tabular}
\end{table}

\paragraph{Results.} \Cref{tab:single} gives the overlaps.
Paired by question, the single-rollout cruxes' gap is 20.4 points smaller than the a priori cruxes' in March Madness ($p = 3 \times 10^{-15}$) and 26.3 points smaller in Metaculus ($p = 6 \times 10^{-5}$).
At $k = 5$, 12 and 25 the single-rollout cruxes' gap on the stripped multiple-rollout cruxes is 0.8, 3.4 and 6.4 points in March Madness and 4.5, $-1.8$ and $-4.5$ in Metaculus, where it is never significant ($p$ from 0.25 to 0.94).
Under the strict and broad levels, on the same stripped multiple-rollout cruxes, the single-rollout cruxes' gap in March Madness is 6.6 ($p = 2 \times 10^{-12}$) and 7.0 ($p = 6 \times 10^{-10}$) points, against 28.1 and 30.6 for the a priori cruxes.
In Metaculus the single-rollout cruxes' gap is $-7.0$ and $-4.8$, above the baseline at the strict level (13 of 15 events, $p = 0.004$ in that direction) and at it at the broad level ($p = 0.94$), against 33.3 and 21.3 for the a priori cruxes.
A separate check extracts every forecast on its own and pools the sets of cruxes of each game or event (9 per game, 29 or 30 per event), which tests the extraction method rather than one forecast.
On the full multiple-rollout cruxes at $k = 25$ that pool overlaps by 28.2\% against 30.8\% in March Madness (gap 2.7, $p = 9 \times 10^{-4}$) and by 60.5\% against 65.0\% in Metaculus ($p = 0.21$).
Extracting one forecast at a time and pooling therefore gives cruxes close to those extracted from pairs of forecasts, though 2.7 points short of the baseline in March Madness.
The tests count residues of $10^{-9}$ or less as ties (Metaculus at $k = 5$, for example, is 10 / 1 / 4 with $p = 0.001$).

\paragraph{What the overlap does not see.} That the single-rollout cruxes sit at the baseline means they are exchangeable with the multiple-rollout cruxes at the resolution of the clustering, not that one forecast is as good as the pool.
The overlap never sees how often a crux recurs across forecasts, whether a crux separates forecasts that disagree, or how much of the pool a set of cruxes covers, since both sets are cut to $k$.
Pooling many forecasts buys those three, and none of them enters the overlap.
The containment check below measures the third, and in March Madness the single-rollout cruxes fall short on it.

\paragraph{Containment.} Overlap counts clusters and is symmetric, so it says nothing about direction.
Containment asks two directional questions on the same clustering: the share of set A's cruxes whose cluster holds at least one crux of set B (is what A says already in the pool), and the share of B's cruxes whose cluster holds at least one crux of A (how much of the pool A covers).
The baseline for each comes from the same label shuffle with the clustering fixed, so it is high where clusters are large (80\% to 95\% in Metaculus at $k = 25$) and depends on how many cruxes carry the A label, which is why a 12-crux forecast and 25 a priori cruxes have different baselines on the same pool.
The first direction rewards sets whose cruxes lump together: in 12 March Madness games 21 to 25 of the 25 a priori cruxes sit in one to three mixed clusters, so their share-in-pool is above baseline while their coverage is below baseline in all 12.
Coverage is therefore the safer direction, and the split-half check sits at baseline on share-in-pool and falls 2.6 points short on coverage at $k = 25$ ($p = 0.035$, \cref{tab:containment}).
In Metaculus it sits at baseline in both directions, 96.0\% against 96.5\% and 95.5\% against 96.5\%.
The baseline is the expectation for a random draw of the same length, not a ceiling: a set of 12 cruxes can cover at most the multiple-rollout cruxes in the 12 largest pool clusters, on average 90\% of the 25 in March Madness and below 100\% in 37 of 59 games, so a chance-corrected score, observed minus baseline over ceiling minus baseline, would read 0 at the baseline and 1 at that ceiling; the paper reports the raw gap and leaves that normalization to future work.

\begin{table}[t]
\caption{Predicted cruxes cover far less of the discovered cruxes than the permutation baseline would, and in March Madness cruxes from one forecast fall short too.
Containment at $k = 25$, same resolution, each direction against its own baseline.
In pool is the share of the set's cruxes whose cluster holds a pooled crux, covers pool the share of the pooled cruxes whose cluster holds one of the set's; gaps are in points, $p$ is one-sided exact Wilcoxon, and MM is March Madness.}
\label{tab:containment}
\centering
\small
\setlength{\tabcolsep}{3.5pt}\begin{tabular}{@{}llrrrlrrrl@{}}
\toprule
& & \multicolumn{4}{c}{In pool} & \multicolumn{4}{c}{Covers pool} \\
\cmidrule(lr){3-6}\cmidrule(lr){7-10}
Source & Cruxes, pool & obs & base & gap & $p$ & obs & base & gap & $p$ \\
\midrule
MM & a priori, headline & 32.7 & 65.3 & 32.5 & $10^{-12}$ & 23.2 & 65.3 & 42.1 & $2 \times 10^{-18}$ \\
MM & single-rollout, stripped pool & 36.7 & 57.7 & 21.0 & $3 \times 10^{-14}$ & 28.6 & 37.8 & 9.2 & $8 \times 10^{-8}$ \\
MM & a priori, stripped pool & 41.1 & 68.3 & 27.3 & $10^{-9}$ & 25.7 & 68.4 & 42.7 & $2 \times 10^{-18}$ \\
MM & per-forecast pooled, headline & 40.9 & 52.2 & 11.3 & $3 \times 10^{-12}$ & 55.0 & 52.1 & $-2.9$ & 0.95 \\
MM & split half & 68.8 & 68.7 & $-0.1$ & 0.73 & 65.9 & 68.5 & 2.6 & 0.035 \\
Met. & a priori, headline & 76.8 & 95.1 & 18.3 & 0.0013 & 75.2 & 95.2 & 20.0 & 0.001 \\
Met. & single-rollout, stripped pool & 87.9 & 90.9 & 3.0 & 0.23 & 82.9 & 79.6 & $-3.3$ & 0.85 \\
Met. & a priori, stripped pool & 75.5 & 93.6 & 18.1 & 0.013 & 80.3 & 93.7 & 13.4 & $3 \times 10^{-4}$ \\
Met. & per-forecast pooled, headline & 89.1 & 95.0 & 5.9 & 0.032 & 98.1 & 95.0 & $-3.1$ & 1 \\
\bottomrule
\end{tabular}
\end{table}

\Cref{tab:containment} gives both directions.
The a priori cruxes are below baseline in both directions in both sources, and the same holds under the strict and broad levels on the full multiple-rollout cruxes.
In March Madness the single-rollout cruxes are also below baseline in both directions, by 21 points on share-in-pool and 9 on coverage; in Metaculus they sit at baseline in both, with baselines of 90.9\% and 79.6\% and 15 events leaving little room to detect anything.
On the same stripped multiple-rollout cruxes the a priori cruxes cover far less of the pool than the single-rollout cruxes do: coverage gaps of 42.7 against 9.2, the a priori cruxes further below the baseline in 55 of 59 games ($p = 10^{-15}$), and 13.4 against $-3.3$ in Metaculus (13 of 15, $p = 0.001$).
On share-in-pool the two sets are separated at $k = 25$ (one-sided $p = 0.038$ and 0.032) and at $k = 12$.
Containment, like overlap, never sees frequency or contrast, and share-in-pool for a short set is measured against a 25-crux slice of the pool, not the whole pool.

\subsection{Open-weight replication protocol}\label{app:openweight}

The open-weight models, whose weights are public and which run on local hardware, are qwen3:32b, gemma4:31b and gpt-oss:120b, named by their tags in the software that served them; the last of them comes from OpenAI, like the GPT models.
In each of these runs one model fills the forecaster, predictor and extractor roles, under the same prompts and scoring as the GPT runs.
Mixing the GPT and open-weight models also lets each role go to a different model (below).
Unlike the main GPT runs, the open-weight runs were first scored with repeated cruxes kept (\cref{sec:exp_headline}); the gpt-oss:120b and gemma4:31b rows of the table are the rescoring without them, described below.

The qwen3:32b repeat follows the March Madness protocol, with 5 prediction calls per game and no filter on either set (\cref{sec:exp_models}); its sets hold no repeated cruxes.
The gpt-oss:120b repeat departs from it once.
Its discovered cruxes pass through a structural filter, which removes the 6.9\% of them (131 of 1,905) that carry decoding debris, such as a question cut short or an option that is a stray token.
An earlier gpt-oss:120b run made 15 prediction calls per game, capped at 30 cruxes per call, with up to 3 attempts, removal of repeated cruxes and a filter on the form of the cruxes in both sets.
The numbers reported in \cref{sec:res_models} come instead from a rebuild at the paper's own 5 unfiltered prediction calls.
With repeated cruxes removed, the rebuild's gaps of 7.1 points at $k = 12$ and 12.7 at $k = 25$ sit within two points of the 15-call run's 8.0 and 14.4, inside the noise of a redraw (below).
The gemma4:31b repeat follows the earlier gpt-oss:120b protocol, with 15 prediction calls per game, at most 30 cruxes per call, up to 3 attempts and the structural filter on both sets, which removes 2 of its 1{,}375 discovered cruxes and 35 of its 6{,}279 a priori cruxes.
The model gpt-oss:120b writes more specific cruxes than gpt-5.4-nano-2026-03-17 and qwen3:32b.
Counted with repeated cruxes kept, 54\% of its cruxes name an explicit numeric threshold, against about 6\% for gpt-5.4-nano-2026-03-17 and qwen3:32b, so they barely cluster even with each other.
Few clusters can then be shared at all, and the baseline itself is low (\cref{tab:openweight}).

Removing repeated cruxes changes little.
No set of a priori cruxes in any run holds a repeat; the repeats are all among the multiple-rollout cruxes, 6 copies across 5 games for gpt-oss:120b, 28 across 14 for gemma4:31b, and none for qwen3:32b.
Rescored without them, the gaps move by at most 1.1 points: gemma4:31b from 15.3 to 15.9 at $k = 12$ (55 games below their baseline, 2 above, 2 equal, $p = 4 \times 10^{-16}$) and from 19.8 to 19.7 at $k = 25$; gpt-oss:120b from 8.5 to 8.0 and from 14.6 to 14.4 in the 15-call run, and from 7.1 to 7.1 at $k = 12$ in the rebuild (48 below, 10 above, 1 equal).
No direction or significance changes, and the cross-model cells below are reported after the same removal.

\begin{table}[t]
\caption{For each of three open-weight models, predicted cruxes match discovered ones less often than the permutation baseline in most games.
The runs repeat the March Madness experiment at the same-resolution level; the qwen3:32b row holds no repeated cruxes, and the gpt-oss:120b and gemma4:31b rows are rescored with repeated cruxes removed.
Gaps are in points, and the counts are games below, above and equal to their own baseline, where equal means that the game's overlap matches the mean of its shuffles.
The model gpt-oss:120b is quoted at $k = 12$ because its extraction yields fewer than 25 cruxes in 17 games.
The model gemma4:31b is also quoted at $k = 12$, the largest length at which most of its games are full, since its extraction yields fewer than 12 cruxes in 17 games.
A game with fewer than $k$ cruxes on one side is scored on all the cruxes it has (\cref{sec:exp_headline}).}
\label{tab:openweight}
\begin{center}
\small
\small\begin{tabular}{@{}lcccccc@{}}
\toprule
Model & $k$ & Overlap (\%) & Baseline (\%) & Gap & Below / above / equal & $p$ \\
\midrule
qwen3:32b & 25 & $38.3 \pm 2.2$ & 52.3 & $13.9 \pm 2.0$ & 39 / 5 / 15 & $1.4 \times 10^{-9}$ \\
gpt-oss:120b & 12 & 11.5 & 18.6 & 7.1 & 48 / 10 / 1 & $2 \times 10^{-10}$ \\
gemma4:31b & 12 & $14.4 \pm 1.8$ & 30.3 & $15.9 \pm 1.3$ & 55 / 2 / 2 & $4 \times 10^{-16}$ \\
\bottomrule
\end{tabular}
\end{center}
\end{table}

A second draw of the a priori cruxes of gpt-5.4-nano-2026-03-17 shows how far a run can move from the draw of the cruxes alone.
At 25 cruxes and the same-resolution level the second draw moves the headline gap by 0.3 points once repeated cruxes are removed (2.1 points before).
Over all levels and set sizes the largest move is 2.2 points for gpt-5.4-nano-2026-03-17 and 2.1 for a redraw of qwen3:32b's cruxes.
Differences between runs of about two points or less can therefore come from the draw of the cruxes alone.
Each repeat is scored against its own baseline, because the baseline depends on how many cruxes each set holds and how they cluster, and both change with the model.
The baseline does not absorb every effect of the model, because an extractor that writes fewer cruxes leaves more room for an a priori crux to match one (below).

\paragraph{Comparing the rows.}
The rows are quoted at different set sizes, and the overlap and its baseline both change with set size (\cref{sec:res_headline}), so the table shows the direction of the gap for each model but does not compare its size across models.

\paragraph{Varying the roles.}\label{app:crossmodel}
The March Madness design uses one model to forecast, to predict cruxes in advance, and to extract them from the forecasts.
The open-weight runs let the roles be separated, and this paragraph records what that shows, though the paper's claim does not rest on it.

Three designs separate the roles, and each answers one question.
Two forecaster-predictor pairs by two extractors ask whether the test can detect a change of model at all.
A grid of three predictors by three extractors asks whether any pairing of those two roles closes the gap below the baseline.
A $2 \times 2 \times 2$ cross of forecaster, predictor and extractor asks whether a model predicts the cruxes of its own rollouts better than those of another model's rollouts.
If it does no better on its own, the gap is not specific to a model predicting itself.

The first design is a positive control, a change expected to move the gap.
Each forecaster is paired with its own predictor, for gpt-5.4-nano and qwen3:32b, and only the extractor varies, which gives two pairs by two extractors at three levels, twelve cells.
A different extractor writes a different number of cruxes in different words from the same rollouts, so the multiple-rollout cruxes change and the gap should change with them.
All twelve cells fall below their baselines, by 8.4 to 19.4 points at $k = 12$ and 12.2 to 28.3 points at $k = 25$.
Swapping the extractor from gpt-5.4-nano to qwen3:32b on the same forecasts shrinks the gap by 11.15 points at $k = 25$ ($p = 9.6 \times 10^{-10}$, paired over 59 games).
From the same forecasts gpt-5.4-nano writes about 2.7 times as many cruxes per game as qwen3:32b, 160 cruxes against 59.
The gap moves under this swap, so the scoring can register a change of about 11 points.
Comparing the two forecaster-predictor pairs moves the gap by 0.70 points ($p = 0.64$), but that comparison swaps the predictor along with the forecaster.

The second design, a grid of three predictors against three extractors at all three levels, gives 27 cells, all below their baselines.
They fall below by 7.7 to 21.5 points at $k = 12$ and by 12.2 to 32.0 points at $k = 25$.

The third design, in which forecaster, predictor and extractor are each gpt-5.4-nano or qwen3:32b, can change the forecaster while the predictor stays fixed.
Swapping the forecaster that way moves the gap by 2 to 3 points at $k = 25$, so the forecasting model is not wholly without effect.
The same design separates two explanations of any advantage a model has on its own forecasts.
One is self-knowledge, an advantage when the predictor is also the forecaster.
The other is idiom, an advantage when the predictor is also the extractor.
One model then wrote both sets of cruxes in the same idiom, and the reranker matches a shared idiom more readily.
The advantage divides about evenly between them, roughly a point each on average.
Self-knowledge reaches significance in two of the six combinations of level and $k$ and idiom in one, against an extractor effect of 6 to 14 points in the same design.
A model is therefore at most marginally more accurate about its own forecasts than about another model's, and the failure reported here is not specific to a model reading itself.

\subsection{Regenerating the predicted cruxes}\label{app:regeneration}

An earlier version of the prediction prompt asked for the parts of each crux in three separate sequences whose entries were meant to line up by position.
Where the entries did not line up, parts of different cruxes were joined, corrupting 13.3\% of the cruxes that prompt produced.
The reported March Madness numbers on gpt-5.4-nano use cruxes regenerated as one JSON object each, none of them malformed.
The Metaculus cruxes were generated once, on 2026-03-31, in the three-sequence format and never regenerated.
A structural check flags none of the 4,727 Metaculus cruxes.
Against the corrupted set, observed overlap is unchanged within noise and the gap below the baseline is 3 to 5 points smaller, so the defect did not produce the finding.

\section{Models and Costs}\label{app:models_costs}

\subsection{Models behind each role}\label{app:roles}

\Cref{tab:roles} lists the model behind each role in each source.

\paragraph{Metaculus events and March Madness reports.}
The 15 events are all the distinct events in the system's forecast data.
Each rollout is one full run of that system, which combines several estimates into one probability (\cref{app:metaculus}).
\Cref{app:reports} reproduces the three reports for Iowa State.
In that example, the analyst report recounts the team's season through its players and recent games, the statistician report lists the team's national rankings and its scoring, shooting and turnover figures, and the matchup-analyst report describes its style of play and how an opponent could beat it.
A turnover is a possession lost before the team can shoot.
In March Madness, gpt-5.4-nano fills all three roles.
Three open-weight models, qwen3:32b, gemma4:31b and gpt-oss:120b, each fill all three roles in a further March Madness run (\cref{sec:exp_models}).
In Metaculus each role goes to a different model.
The predictor is gpt-5.4, which costs about 40 times as much per token as the forecaster at OpenAI's list prices.
Price is a rough guide to capability, so in Metaculus the predictor is unlikely to be weaker than the forecaster.

\paragraph{Matching.}
The same reranker (\cref{sec:equivalence}) decides whether two cruxes match in every experiment (\cref{tab:roles}).

\begin{table}[t]
\caption{In March Madness one model forecasts, predicts and extracts, while in Metaculus each role goes to a different model.
The model behind each role, behind the March Madness scouting reports and behind the matching of cruxes.}
\label{tab:roles}
\begin{center}
\small
\small\begin{tabular}{@{}lll@{}}
\toprule
 & March Madness & Metaculus \\
\midrule
Forecaster & gpt-5.4-nano & gpt-5-nano, with web search \\
Predictor & gpt-5.4-nano-2026-03-17 & gpt-5.4 \\
Extractor & gpt-5.4-nano & gpt-5.4-mini \\
\midrule
Scouting reports & gpt-5.4-mini, with web search & none \\
Matching & Qwen3-Reranker-8B & Qwen3-Reranker-8B \\
\bottomrule
\end{tabular}
\end{center}
\end{table}

\subsection{Computational costs}\label{app:costs}

\Cref{tab:costs} summarizes API and compute costs for all experiments.
Three March 2026 stages were not metered per stage, so their cells read ``not recorded''.
The month's total OpenAI bill for all projects, including development iterations and unrelated experiments, was \$2{,}314.

Matching ran on the university cluster's unmetered GPUs.
The headline scored 271{,}950 pairs in 1{,}087{,}800 reranker passes, 6.4 GPU-hours of scoring and 57 minutes of wall clock, at no dollar cost.
Clustering, the cut and the permutation baseline call no model and take under a millisecond per game on a CPU.
Judging the same pairs with gpt-5.4-nano-2026-03-17 instead would cost an estimated \$97 to \$109 at standard rates, or \$49 to \$55 at batch rates, and a local qwen3:32b judge an estimated 550 GPU-hours, 86 times the reranker.
Those are token counts priced at list rates, not billed amounts.
Per game, at batch list prices without the cached-input discount, the five a priori calls cost about \$0.012, the single-rollout extraction \$0.0013, and the 36 multiple-rollout comparisons an estimated \$0.06 to \$0.08.
The rollouts themselves were never metered.

\begin{table}[H]
\caption{API and compute costs for all experiments.
Costs are billed amounts where a record exists, dashboard readings for the March 2026 runs and token counts at OpenAI list prices of September 2026 for the later batch runs, and estimates where a figure is given without a record; cells with no per-stage record read not recorded, and batch rates apply to the runs marked batch, at half the standard rates (OpenAI list prices, 2026-09-22).
Where the model column names a snapshot, that is the snapshot the run record names; elsewhere the column gives the undated model name the calls were sent with.
Which teams meet in later rounds depends on who wins the earlier games, so March Madness forecasts were made for every pairing of the tournament's teams.
None of the 59 games draws on the forecasts for the First Four, the tournament's play-in round.}
\label{tab:costs}
\begin{center}
\footnotesize
\setlength{\tabcolsep}{4pt}
\small\begin{tabular}{@{}p{4.8cm}lrc@{}}
\toprule
Experiment & Model & Calls & Cost \\
\midrule
\multicolumn{4}{@{}l}{\textit{March Madness}} \\
Scouting reports & \begin{tabular}[t]{@{}l@{}}gpt-5.4-mini\\ + web search\end{tabular} & not recorded & not recorded \\
Forecasts, 62 teams & gpt-5.4-nano (batch) & 17{,}019 stored & not recorded \\
Forecasts, First Four & gpt-5.4 (batch) & 3,483 & $\sim$\$100 \\
Multiple-rollout extraction & gpt-5.4-nano & 68,076 & $\sim$\$290 \\
A priori prediction & gpt-5.4-nano & 2{,}341 & not recorded \\
Single-rollout extraction, 531 forecasts & gpt-5.4-nano-2026-03-17 (batch) & 531 & \$0.69 \\
A priori prediction, regenerated, 59 games, 2 samples & gpt-5.4-nano-2026-03-17 & 590 & $\sim$\$2 \\
A priori prediction, same-day control, 59 games & gpt-5.4-nano-2026-03-17 (batch) & 295 & \$0.40 \\
Specific-prompt arms, 59 games (specific, exemplar and ladder) & gpt-5.4-nano-2026-03-17 (batch) & 885 & \$2.29 \\
Open-weight replication, local GPUs & \begin{tabular}[t]{@{}l@{}}qwen3:32b, gemma4:31b,\\ gpt-oss:120b\end{tabular} & not recorded & \$0 \\
Cross-model check (March Madness, two predictors by two extractors) & gpt-5.4-nano-2026-03-17 (batch) & 2{,}124 & \$3.74 \\
Evidence statements for evidence matching, 59 games & gpt-5.4-nano-2026-03-17 (batch) & not recorded & \$5.10 \\
Saturation scoring, every pair of the multiple-rollout cruxes & \begin{tabular}[t]{@{}l@{}}Qwen3-Reranker-8B\\ (local, 12 GPU-hours)\end{tabular} & 751{,}515 pairs & \$0 \\
\midrule
\multicolumn{4}{@{}l}{\textit{Metaculus}} \\
Forecasts & \begin{tabular}[t]{@{}l@{}}gpt-5-nano\\ + web search\end{tabular} & not recorded & not recorded \\
Multiple-rollout extraction & gpt-5.4-mini & 6,467 & $\sim$\$62 \\
Single-rollout extraction, 448 forecasts & gpt-5.4-mini-2026-03-17 (batch) & 448 & \$1.37 \\
A priori prediction & gpt-5.4 & 825 & \$11.66 \\
Specific-prompt arms, 15 events (three prompts) & gpt-5.4 & $\sim$2{,}475 & $\sim$\$21 \\
\midrule
\multicolumn{4}{@{}l}{\textit{Both sources}} \\
GPU scoring & \begin{tabular}[t]{@{}l@{}}Qwen3-Reranker-8B\\ (local A100, H200)\end{tabular} & not recorded & \$0 \\
Specificity tags and coarsening rewrites & qwen3:32b (local) & not recorded & \$0 \\
\bottomrule
\end{tabular}
\end{center}
\end{table}

\section{Data and Prompts}\label{app:data_prompts}

\subsection{Metaculus data}\label{app:metaculus}

The Metaculus forecasts come from an automated forecasting system built by one of the authors for other work, run on gpt-5-nano with web search.
The system searches for existing forecasts and for factors bearing on the event, writes three probability estimates from three prompts (an outside view, an inside view and a contrarian view), and combines them by a log-odds mean; a critique pass then challenges that mean, and a calibration call that sees both writes the final probability; the extractor reads the rendered forecast text, not the three estimates, the mean or the critique.
Each rollout runs its own research phase: three search-enabled model calls write its forecasts brief, from a consensus, a contrarian and a recent-news frame, and three more write its causal brief, each a separate API call with the web search tool, so the model searches live inside each call and no search result is shared across rollouts; the three persona estimates then read the briefs and search nothing.
Thirty rollouts were generated per event and 449 of the 450 were stored, leaving 29 or 30 per event.
The forecast data contained 16 entries, two of them for the same event, the Christie's/Sotheby's auction record, and both entries were kept, so that event's forecasts come from both.
\Cref{tab:events} lists the 15 events with their Metaculus post identifiers; three identifiers are ambiguous because two posts carry the same question, and the table gives the post in the Spring 2026 FutureEval Bot Tournament, where the other twelve sit.
At the grain of one label per event the events span seven domains, eight with March Madness's basketball.

\begin{table}[t]
\caption{The 15 Metaculus events, their post identifiers and one domain label each.}
\label{tab:events}
\begin{center}
\small
\small\begin{tabular}{@{}lp{7.9cm}l@{}}
\toprule
Post & Question & Domain \\
\midrule
42117 & Will AfD win the most seats in the Baden-W\"urttemberg Landtag election? & elections \\
42116 & Will CDU win the most seats in the Baden-W\"urttemberg Landtag election? & elections \\
42231 (or 40005) & Will any political party or coalition acquire a supermajority in the 2026 Hungarian parliamentary elections? & elections \\
42118 & Will the official result of the Lamag by-election (Sabah), scheduled for January 24, 2026, be overturned before May 1, 2026? & elections \\
42232 (or 42148) & Will Ali Khamenei cease to be supreme leader of Iran before April 1, 2026? & geopolitics \\
42119 & Will any state actor use chemical weapons in Ukraine before May 1, 2026? & geopolitics \\
42246 & Will H.R. 33 (United States--Taiwan Expedited Double-Tax Relief Act) be enacted into law before May 1, 2026? & law and policy \\
42230 (or 42135) & Will the UK increase the qualifying period for settlement to 10 years before May 2026? & law and policy \\
42245 & Will the IOM Missing Migrants Project record at least 1,000 Mediterranean migrant deaths in 2026 before May 2026? & migration \\
42237 & Will the IOM Missing Migrants Project record at least 900 Mediterranean migrant deaths in 2026 before April 2026? & migration \\
42120 & Will Frontier Group Holdings, Inc.\ file for bankruptcy protection before May 1, 2026? & business and markets \\
42243 & Will Christie's, Sotheby's, or Phillips report a public auction sale of a painting for at least US\$100 million before May 1, 2026? & business and markets \\
42241 & Will at least one film surpass US\$1 billion in worldwide box office before May 1, 2026? & culture \\
42240 & Will the Cannes Film Festival publish its 2026 official selection before April 10, 2026? & culture \\
42235 & Will the English Wikipedia have at least 7,145,000 articles before March 1, 2026? & media and internet \\
\bottomrule
\end{tabular}
\end{center}
\end{table}

\subsection{Prompts}\label{app:prompts}

This appendix gives every prompt verbatim: the eleven prediction strategies used on Metaculus questions, the five prediction variants used on March Madness matchups, the three extraction prompts (two pairwise and one single-forecast), and the three scouting-report persona prompts.
All eleven feed the a priori cruxes, a priori in this paper's sense of reading no rollout.
The names direct\_zero\_shot and few\_shot\_exemplar use shot in the prompting sense of worked examples.
Every call ran at the API default temperature with no other sampling overrides.
In the first Metaculus prompt, the runs of dashes that only separate its sections are shortened to fit the column.

\raggedcolumns
\begin{multicols}{2}

\subsubsection{Prediction prompts for Metaculus (ids 0--10)}

\paragraph{id=0 (clone\_introspection\_baseline)}
\begin{Verbatim}[breaklines=true, breakafter=-, fontsize=\footnotesize, frame=single, framerule=0.3pt]
Consider the following forecasting question. What do you predict would be your own cruxes when forecasting this question?

To clarify:
- I will provide you with a forecasting question.
- Imagine what many independent copies of you might respond when asked to forecast this question; consider the variety of probability estimates, reasoning processes, evidence cited, and assumptions made. Note that this is more narrow than simply imagining all possible forecasts; it's a matter of introspection about how you think you (and hence your clones) would *likely* respond if asked multiple times independently. So the trick is to consider only the *likely subset* of all possible variations.
- Then, use the definition of cruxes provided below, to predict the distributional cruxes between those clones' forecasts.

---------- FORECASTING QUESTION ----------
{question}

---------- DEFINITION OF DISTRIBUTIONAL CRUXES ----------

In order to build up to a notion of cruxes across a *distribution* of forecasts (i.e., many independent copies of you forecasting the question above), we first need to define cruxes between *pairs* of forecasts. Consider this to be an extended thought experiment, which implicitly defines what we actually care about (i.e., distributional cruxes) by first defining pairwise cruxes.

DEFINING "PAIRWISE" CRUXES BETWEEN TWO FORECASTS:

A *crux* between two forecasts is a question whose answer could change the overall conclusion of the forecasts. Cruxes arise when two forecasts differ in their reasoning steps, evidence cited, assumptions made, or intermediate conclusions reached -- even if they ultimately arrive at similar probabilities.

That is, one might extract a list of crux questions between two forecasts by following Algorithm 1 below:

ALGORITHM 1: EXTRACT PAIRWISE CRUX QUESTIONS BETWEEN TWO FORECASTS:
Hypothetical Inputs:
    - The forecasting question.
    - FORECAST A: The first forecast to analyze.
    - FORECAST B: The second forecast to analyze.
1. Systematically analyze both forecasts:
   For FORECAST A, identify:
   - Key factual claims (what world-facts does it treat as true?)
   - Reasoning steps (how does it proceed logically?)
   - Underlying assumptions (what unstated premises about the world does it rely on?)
   - Intermediate conclusions (what sub-conclusions does it reach?)
   For FORECAST B, do the same analysis.
   Then, for each dimension above, explicitly note whether the forecasts differ.
2. Distill any differences into cruxes -- questions about objective facts of the world.
   - A crux must be about an objective state of the world -- something that could in principle be resolved by examining reality, not by reading either forecast.
   - Do NOT frame cruxes as beliefs, assumptions, or claims of either forecast.
     Bad:  "Does the forecast assume the Fed will raise rates?"
     Good: "Will the Fed raise rates in the relevant period?"
   - Avoid using the probability estimate itself as a crux.
   - Write each crux as a neutral question (e.g., "Is X true?").
   - Do NOT mention "Forecast A" or "Forecast B" in any crux question.
   - Return at least one crux if the forecasts rely on different world-facts -- even if they reach similar probabilities.
   - Return no cruxes ONLY if the forecasts are word-for-word identical or trivial rephrasings with zero substantive difference in reasoning.
   - Identify one crux per genuinely distinct point of disagreement -- don't split a single difference into multiple cruxes, and don't merge separate differences into one.
3. For each crux, record each forecast's answer as a standalone premise.
   - Example: If the crux is "Is X true?" then the premises might be:
       - Forecast A -> "X is true."
       - Forecast B -> "X is not true."
   - Do NOT mention "Forecast A" or "Forecast B" in any premise.
   - Each premise must be an objective claim about the world, not a statement about what a forecast says or believes.
   - The two premises must contradict each other along the dimension the crux asks about.
END ALGORITHM 1

Using this algorithm, one can extract crux questions from any pair of forecasts. Note that even if the forecasts reach similar probabilities, the algorithm will still produce cruxes if they differ in reasoning, evidence, assumptions, or intermediate conclusions.

Furthermore, each crux question is binary (i.e., has exactly two possible answers, arbitrarily called "A" and "B", corresponding to each forecast's premise), and these premises directly contradict each other.

CONVERTING A SET OF PAIRWISE CRUXES TO DISTRIBUTIONAL CRUXES:
Given a set of multiple forecasts (e.g., from many independent copies of you), one could imagine aggregating all the pairwise crux questions to identify the main axes of disagreement across the *entire* set of forecasts. And, assuming that set were arbitrarily large, this would converge to a distributional notion of cruxes!

This could look something like the following Algorithm 2:
ALGORITHM 2: AGGREGATE PAIRWISE CRUX QUESTIONS ACROSS MULTIPLE FORECASTS:
Hypothetical Inputs:
    - the unioned list of **pairwise crux questions** from all forecast pairs.
1. Cluster these pairwise crux questions based on semantic similarity and the dimensions of disagreement they represent.
  - Given k crux questions which are all getting at the same core difference, group them together.
2. Identify the subset of forecasts that each **cluster** of crux questions applies to.
3. Choose one crux question to represent the entire cluster.
  - This ensures the representative crux question has the same structure: it remains a binary question whose answers contradict each other.
4. Assign the representative crux question the cluster's subset of forecasts: note which forecasts provide the "A" premise and which provide the "B" premise.
END ALGORITHM 2

Of course, this isn't an explicit definition of distributional cruxes, but you can get a gist of the idea. The end goal is to get a generic list of cruxes which capture the factors of variation from the *distribution* of your clones' forecasts.

---------- RESTATEMENT OF OVERALL TASK ----------

Your task is to imagine what many independent copies of you would forecast for the question above, well enough that you can predict the key cruxes between those forecasts. Then, using the implicit notion of distributional cruxes described above, predict the most important cruxes between those forecasts.

1. First restate the overall task in your own words.
2. Then, suggest a rough imagining of the variety of probability estimates, reasoning processes, evidence cited, assumptions, and conclusions your clones would produce.
3. Then jump directly to guessing the distributional cruxes between those imagined forecasts.
   - Formatting reminders:
     - A crux is a question whose answer could change the overall conclusion.
     - Focus on the underlying reasoning differences, not the probability estimate itself.
     - Write each crux as a neutral question (e.g., "Is X true?").
     - For the final JSON output: group the conflicting premises into two arbitrary opposing sides (Perspective A and Perspective B).
     - Prioritize the most important cruxes that capture the main axes of disagreement.

tl;dr: Imagine the variety of likely forecasts your clones would produce for the question above, then identify the cruxes that capture the main axes of disagreement across those (hypothetical) forecasts.
\end{Verbatim}

\paragraph{id=1 (direct\_zero\_shot)}
\begin{Verbatim}[breaklines=true, fontsize=\footnotesize, frame=single, framerule=0.3pt]
Forecasting question:
{question}

What are the key cruxes for this forecasting question? A crux is a factual question where disagreement between forecasters would lead to different probability estimates.

For each crux:
- State it as a neutral question (e.g., "Is X true?").
- Provide two contradictory premises (A and B) representing the two sides.

Also briefly restate what you understand the task to be, and sketch the range of forecasts that reasonable people might give.
\end{Verbatim}

\paragraph{id=2 (expert\_forecaster\_persona)}
\begin{Verbatim}[breaklines=true, fontsize=\footnotesize, frame=single, framerule=0.3pt]
You are a superforecaster with decades of experience in geopolitical, economic, and scientific forecasting. You have been trained at the Good Judgment Project, have a published track record of well-calibrated predictions, and routinely mentor junior forecasters.

A junior forecaster comes to you with the following question and asks: "What are the key points where smart forecasters would disagree on this?"

Forecasting question:
{question}

Draw on your extensive experience to identify the cruxes -- the factual questions where reasonable, well-informed forecasters would genuinely disagree, and where that disagreement would drive different probability estimates.

For your response:
1. Briefly restate what the junior forecaster is really asking (task_recap).
2. Describe the range of forecasts you'd expect from a room of superforecasters -- what probability ranges, what reasoning approaches, what evidence they'd weight differently (distribution_sketch).
3. Identify the crux questions -- each as a neutral factual question (e.g., "Will X happen?") with two contradictory premises (A and B) representing the opposing positions forecasters would take.

Focus on cruxes that are genuinely uncertain and decision-relevant, not trivial or obvious points.
\end{Verbatim}

\paragraph{id=3 (devils\_advocate)}
\begin{Verbatim}[breaklines=true, fontsize=\footnotesize, frame=single, framerule=0.3pt]
Consider this forecasting question:
{question}

Your task has three phases:

PHASE 1 -- ARGUE FOR A HIGH PROBABILITY:
Make the strongest possible case that the answer to this question is "yes" (or that the probability should be high). Marshal every relevant fact, trend, analogy, and argument that supports this position. Be specific and forceful.

PHASE 2 -- ARGUE FOR A LOW PROBABILITY:
Now reverse course entirely. Make the strongest possible case that the answer is "no" (or that the probability should be low). Attack every argument you just made. Find counterevidence, alternative explanations, and reasons to be skeptical.

PHASE 3 -- IDENTIFY DIVERGENCE POINTS:
Now step back. Where exactly do these two arguments diverge? What are the specific factual questions where the two sides rely on different assumptions about the world? These are the cruxes.

For each crux:
- State it as a neutral question (e.g., "Is X true?").
- Give premise A (what the high-probability side assumes) and premise B (what the low-probability side assumes). These must directly contradict each other.

Also provide a brief task recap and a sketch of the distribution of forecasts reasonable people might give.

Important: Cruxes should be about objective facts, not about the forecasting question itself. Do not use the conclusion as a crux.
\end{Verbatim}

\columnbreak
\paragraph{id=4 (chain\_of\_thought)}
\begin{Verbatim}[breaklines=true, fontsize=\footnotesize, frame=single, framerule=0.3pt]
Think step by step about the following forecasting question:
{question}

Step 1: Restate the question and what it's really asking.

Step 2: List the key factors that would influence the outcome. For each factor, note whether its current state or future trajectory is certain or uncertain.

Step 3: For each uncertain factor, consider: would reasonable, well-informed forecasters disagree about this? If so, how? What would one side believe versus the other?

Step 4: Among these disagreements, which ones would most strongly affect the final probability estimate? These are the cruxes.

Step 5: For each crux, formulate it as a neutral factual question (e.g., "Is X true?" or "Will Y happen?"). Then state the two contradictory premises:
- Premise A: What one group of forecasters would believe.
- Premise B: What the opposing group would believe.

Also sketch the range of probability estimates and reasoning that different forecasters would plausibly offer.

Remember: cruxes are about objective facts, not about methodology or framing. Each pair of premises must directly contradict each other.
\end{Verbatim}

\paragraph{id=5 (few\_shot\_exemplar)}
\begin{Verbatim}[breaklines=true, fontsize=\footnotesize, frame=single, framerule=0.3pt]
Your task is to identify the key cruxes for a forecasting question. A crux is a factual question where forecasters would disagree, and where that disagreement drives different probability estimates.

Here are two worked examples:

--- EXAMPLE 1 ---
Question: "Will the European Central Bank raise interest rates by more than 50 basis points at its next meeting?"

Task recap: I need to identify the factual disagreements that would drive different forecasts about the ECB's next rate decision.

Distribution sketch: Some forecasters would estimate 60-70% probability, emphasizing persistent inflation and hawkish rhetoric from ECB officials. Others would estimate 20-35%, pointing to signs of economic slowdown, banking sector stress, and the ECB's historical tendency toward gradualism. A middle group around 40-50% would weigh both factors roughly equally.

Crux questions:
1. "Is eurozone core inflation currently accelerating?"
   A: "Core inflation is accelerating and will remain above 5% through the next meeting."
   B: "Core inflation has peaked and is beginning to decelerate."
2. "Is the eurozone banking sector under significant stress?"
   A: "The banking sector is stable and not a constraint on ECB policy."
   B: "Banking sector stress is significant enough to make the ECB more cautious."
3. "Has the ECB's forward guidance committed it to a large rate hike?"
   A: "Recent ECB communications have effectively pre-committed to a 50+ bp hike."
   B: "The ECB has preserved optionality and could justify a smaller move."

--- EXAMPLE 2 ---
Question: "Will a Category 4 or 5 hurricane make landfall in the continental United States in 2026?"

Task recap: I need to find the factual questions where forecasters would disagree when predicting major hurricane landfall in the US this year.

Distribution sketch: Climate-focused forecasters might estimate 35-45%, citing warmer-than-average sea surface temperatures and active hurricane season forecasts. Historically-minded forecasters might estimate 15-25%, noting that Cat 4/5 US landfalls are relatively rare events even in active seasons. Some might land around 25-35%, combining base rates with current conditions.

Crux questions:
1. "Are sea surface temperatures in the Atlantic Main Development Region anomalously warm for 2026?"
   A: "SSTs are significantly above average, fueling stronger storm development."
   B: "SSTs are near normal or only modestly above average."
2. "Is the current ENSO state favorable for Atlantic hurricane activity?"
   A: "La Nina or neutral conditions prevail, reducing wind shear and favoring Atlantic hurricanes."
   B: "El Nino conditions prevail, increasing wind shear and suppressing Atlantic hurricanes."
3. "Do atmospheric steering patterns favor US landfall this season?"
   A: "Subtropical ridge patterns are likely to steer storms toward the US coast."
   B: "Steering patterns are likely to recurve storms out to sea before US landfall."

--- YOUR TURN ---
Now identify the cruxes for this forecasting question:
{question}

Provide:
1. A brief task recap
2. A sketch of the distribution of likely forecasts
3. The crux questions, each as a neutral question with two contradictory premises (A and B)

Follow the same format as the examples above. Focus on genuinely uncertain factual questions, not the forecasting question itself.
\end{Verbatim}

\paragraph{id=6 (socratic\_questioning)}
\begin{Verbatim}[breaklines=true, fontsize=\footnotesize, frame=single, framerule=0.3pt]
Consider this forecasting question:
{question}

Adopt the Socratic method. Interrogate yourself with probing questions to surface the hidden assumptions, uncertain facts, and contested evidence that underlie any forecast on this topic. Do not accept any claim at face value -- question everything.

Ask yourself:
- What am I assuming about the current state of the world? Is that assumption warranted?
- What evidence would I need to see to dramatically change my probability estimate? Do I have that evidence?
- What do I think I know about this topic that might be wrong?
- Where am I relying on intuition rather than data? Could that intuition be systematically biased?
- What would a smart person who disagrees with me say? Why would they say it?
- Am I anchoring on a particular scenario? What am I neglecting?

After this self-interrogation, identify the cruxes: the specific factual questions where your answers to the above probes would split forecasters into different camps.

For each crux:
- State it as a neutral question (e.g., "Is X true?").
- Provide two contradictory premises (A and B) representing opposing positions.

Also provide:
- A task recap (what you understand you're being asked)
- A distribution sketch (the range of forecasts you'd expect from different people)

Cruxes must be about objective facts, not about beliefs or methodology.
\end{Verbatim}

\paragraph{id=7 (decomposition)}
\begin{Verbatim}[breaklines=true, fontsize=\footnotesize, frame=single, framerule=0.3pt]
Consider this forecasting question:
{question}

Decompose this question into its constituent sub-questions. A good decomposition breaks the overall question into independent (or semi-independent) factual components that, taken together, determine the answer.

For example, a question about whether a company will go bankrupt might decompose into: (a) What is its current debt load? (b) What are its revenue trends? (c) Can it refinance? (d) Is the broader market favorable?

Your steps:
1. Identify 4-8 sub-questions that the main question depends on.
2. For each sub-question, assess: would reasonable forecasters agree or disagree on the answer?
3. The sub-questions where forecasters would disagree are the cruxes.

For each crux:
- State it as a neutral factual question.
- Give premise A (one side's position) and premise B (the opposing position). These must directly contradict each other.

Also provide:
- A task recap showing you understand the question.
- A distribution sketch of the range of forecasts and reasoning you'd expect.

Focus on the most decision-relevant sub-questions -- the ones where disagreement would most shift the probability estimate.
\end{Verbatim}

\paragraph{id=8 (debate\_framing)}
\begin{Verbatim}[breaklines=true, fontsize=\footnotesize, frame=single, framerule=0.3pt]
Two highly intelligent, well-informed forecasters -- call them Forecaster X and Forecaster Y -- disagree about the following question:
{question}

Forecaster X assigns a relatively high probability. Forecaster Y assigns a relatively low probability. They are now debating.

Script their debate. Have them go back and forth for 3-4 exchanges, each time getting more specific about exactly WHERE they disagree. They should be respectful but pointed, drilling down from vague disagreement to specific factual claims about the world.

After scripting the debate, step back and identify the core cruxes -- the specific factual questions that the debate revealed as the real sources of disagreement.

For each crux:
- State it as a neutral question (e.g., "Is X true?").
- Provide premise A (Forecaster X's position) and premise B (Forecaster Y's position). These must directly contradict each other.
- Do NOT mention "Forecaster X" or "Forecaster Y" in the crux questions or premises.

Also provide:
- A task recap
- A distribution sketch of the forecasts you'd expect across many forecasters

Remember: the crux questions and premises must be about objective facts, stated without reference to any forecaster.
\end{Verbatim}

\paragraph{id=9 (causal\_reasoning)}
\begin{Verbatim}[breaklines=true, fontsize=\footnotesize, frame=single, framerule=0.3pt]
Consider this forecasting question:
{question}

Your task is to reason causally about this question.

1. CAUSAL MAP: Identify the key causal factors that influence the outcome. What causes what? Draw out the causal chain from current conditions to the event in question. Include both direct causes and important mediating variables.

2. UNCERTAIN LINKS: For each causal link in your map, assess how certain or uncertain it is. Some causal relationships are well-established; others are speculative, contested, or depend on conditions that are themselves uncertain.

3. CRUX IDENTIFICATION: The most uncertain and most influential causal links are the cruxes. These are the places where forecasters would disagree about how the causal mechanism works or about the state of a key causal variable.

For each crux:
- State it as a neutral factual question about the causal factor or link (e.g., "Does X cause Y in this context?").
- Provide premise A and premise B as contradictory claims about the causal relationship or the state of the causal variable.

Also provide:
- A task recap
- A distribution sketch of the forecasts you'd expect

Focus on causal factors that are both uncertain AND high-leverage -- where getting the causal story wrong would most change the forecast.
\end{Verbatim}

\paragraph{id=10 (pre\_mortem)}
\begin{Verbatim}[breaklines=true, fontsize=\footnotesize, frame=single, framerule=0.3pt]
Consider this forecasting question:
{question}

Perform a pre-mortem analysis.

Imagine it is some time after the resolution date. The consensus forecast turned out to be badly wrong. Now work backwards:

SCENARIO 1: The consensus was too HIGH. The outcome was "no" (or much lower probability than expected). What fact about the world did the consensus get wrong? What assumption turned out to be false? What evidence was ignored or misweighted?

SCENARIO 2: The consensus was too LOW. The outcome was "yes" (or much higher probability than expected). What fact about the world did the consensus underappreciate? What trend was missed? What unlikely event actually happened?

Now, the factual questions that distinguish these failure modes from the "consensus was right" world are exactly the cruxes -- the uncertain facts where forecasters would disagree.

For each crux:
- State it as a neutral question (e.g., "Is X true?").
- Provide premise A and premise B as contradictory factual claims.

Also provide:
- A task recap
- A distribution sketch of the forecasts you'd expect

Focus on cruxes that represent genuine uncertainty, not mere contrarianism. What are the factual questions that, if answered differently, would legitimately shift the forecast?
\end{Verbatim}

\columnbreak
\subsubsection{Prediction prompts for March Madness (ids 20--24)}

\paragraph{id=20 (matchup\_analyst\_grounded)}
\begin{Verbatim}[breaklines=true, fontsize=\footnotesize, frame=single, framerule=0.3pt]
You are analyzing a basketball matchup with detailed scouting reports for both teams.

{question}

## {team_a_name}
{team_a_report}

## {team_b_name}
{team_b_report}

### Task
Many independent analysts have this same evidence. They will disagree about who wins because they weigh uncertain factors differently.

Identify the key factors of disagreement: specific, uncertain world-facts where reasonable analysts would take opposing positions, and where resolving that uncertainty would change the predicted winner.

For each factor:
- State it as a neutral yes/no question about an objective, verifiable fact
- Provide two contradictory positions (A and B), each a specific factual claim
- A and B must be mutually exclusive and jointly exhaustive

Rank by importance -- how much the factor's resolution swings the prediction.

Also provide:
- A task recap (restate what you're doing)
- A distribution sketch (imagine the range of conclusions different analysts would reach)
\end{Verbatim}

\paragraph{id=21 (category\_diversified)}
\begin{Verbatim}[breaklines=true, fontsize=\footnotesize, frame=single, framerule=0.3pt]
You are analyzing a basketball matchup with detailed scouting reports for both teams.

{question}

## {team_a_name}
{team_a_report}

## {team_b_name}
{team_b_report}

### Task
Identify factors where reasonable analysts would disagree about who wins. To ensure breadth, you MUST produce at least one contrast from EACH of these categories:

1. **Statistical matchup**: Shooting efficiency, pace, rebounding differential, turnover rates, defensive metrics
2. **Coaching & strategy**: Scheme adjustments, rotation decisions, timeout patterns, game-plan adaptation
3. **Momentum & psychology**: Recent form, pressure handling, crowd effects, team chemistry
4. **Health & conditioning**: Injuries, rest days, minutes load, depth when starters are in foul trouble
5. **Historical & situational**: Tournament experience, seeding trends, travel/scheduling, past head-to-head
6. **Stylistic clash**: How each team's strengths interact with the other's weaknesses

For each factor:
- State it as a neutral yes/no question about an objective, verifiable fact
- Provide two contradictory positions (A and B), each a specific factual claim
- A and B must be mutually exclusive and jointly exhaustive

Rank by importance -- how much resolving the factor swings the prediction.

Also provide:
- A task recap (restate what you're doing)
- A distribution sketch (imagine the range of conclusions different analysts would reach)
\end{Verbatim}

\paragraph{id=22 (high\_volume\_brainstorm)}
\begin{Verbatim}[breaklines=true, fontsize=\footnotesize, frame=single, framerule=0.3pt]
You are analyzing a basketball matchup with detailed scouting reports.

{question}

## {team_a_name}
{team_a_report}

## {team_b_name}
{team_b_report}

### Task
BRAINSTORM WIDELY. Your goal is to generate as many distinct factors of disagreement as possible -- aim for at least 12. Cast a wide net.

Think about:
- Every uncertain stat that could go either way
- Every coaching decision that's debatable
- Every intangible some analysts weigh heavily and others dismiss
- Every historical pattern that may or may not apply
- Every injury/health concern with uncertain impact
- Every matchup-specific interaction that's hard to predict

For each factor:
- State it as a neutral yes/no question about an objective, verifiable fact
- Provide two contradictory positions (A and B), each a specific factual claim
- A and B must be mutually exclusive and jointly exhaustive

Quantity matters. Include factors you're only 60% sure are important -- downstream filtering will handle quality. Don't self-censor early.

Also provide:
- A task recap
- A distribution sketch
\end{Verbatim}

\paragraph{id=23 (contrarian\_stress\_test)}
\begin{Verbatim}[breaklines=true, fontsize=\footnotesize, frame=single, framerule=0.3pt]
You are analyzing a basketball matchup from a contrarian perspective.

{question}

## {team_a_name}
{team_a_report}

## {team_b_name}
{team_b_report}

### Task
Consider what the consensus prediction would be for this matchup. Now systematically attack it.

1. UPSET FACTORS: What specific, verifiable facts would need to be true for the underdog to win? Matchup advantages, recent trends the market underweights, rest edges, coaching history.

2. FRAGILE ASSUMPTIONS: What does the favorite's case rest on that could break? Star health, shooting sustainability, defensive effectiveness against this specific opponent.

3. HIDDEN VARIABLES: What's hard to assess from scouting reports but could swing the game? Referee tendencies, arena effects, fatigue, emotional factors.

Convert each point of uncertainty into a factor of disagreement:
- State it as a neutral yes/no question about an objective, verifiable fact
- Provide two contradictory positions (A and B), each a specific factual claim
- A and B must be mutually exclusive and jointly exhaustive

Also provide:
- A task recap
- A distribution sketch
\end{Verbatim}

\paragraph{id=24 (multi\_lens\_personas)}
\begin{Verbatim}[breaklines=true, fontsize=\footnotesize, frame=single, framerule=0.3pt]
You are analyzing a basketball matchup through four expert lenses.

{question}

## {team_a_name}
{team_a_report}

## {team_b_name}
{team_b_report}

### Task
Four analysts are debating this matchup. Each sees different things:

**The Statistician**: Efficiency metrics, pace matching, three-point shooting, rebounding rates, turnover differentials, defensive ratings. What statistical factors are genuinely uncertain?

**The Tactician**: Coaching adjustments, defensive schemes, rotation patterns, in-game strategy. What strategic factors could go either way?

**The Scout**: Individual matchups, star player variance, bench depth, injury impacts. What player-level factors are uncertain?

**The Historian**: Tournament experience, seed performance trends, upset patterns, team trajectory. What contextual factors matter?

Each analyst contributes 2-3 factors from their domain. For each:
- State it as a neutral yes/no question about an objective, verifiable fact
- Provide two contradictory positions (A and B), each a specific factual claim
- A and B must be mutually exclusive and jointly exhaustive

Rank all factors by importance across all four lenses.

Also provide:
- A task recap
- A distribution sketch
\end{Verbatim}

\subsubsection{Extraction prompts}

\paragraph{March Madness extraction, contrast\_prompts.json id=0 (structured\_\allowbreak analysis\_\allowbreak baseline)}
\begin{Verbatim}[breaklines=true, fontsize=\footnotesize, frame=single, framerule=0.3pt]
QUALITY RUBRIC -- every crux you produce must satisfy all four criteria:

Binary Contrast: A well-formed crux question poses a clear this-or-that choice. The two options (option_A and option_B) must be mutually exclusive -- choosing one rules out the other.

Intermediate: A good crux addresses an upstream factual sub-question whose answer informs the final forecast. It must NOT be a restatement of the original question or the final answer options. It should identify a specific factual premise that the arguments disagree on, upstream of their conclusions.

World Fact: Both answer options must make objective claims about the state of the world. They must NOT reference a response, forecast, or forecaster (e.g., "Response A believes X"). Each option should be a standalone factual claim.

Clear: The crux must be on-topic, free of internal contradictions, specific enough to evaluate, and easy to understand.

TASK:
1. Systematically analyze both arguments:
   For ARGUMENT A, identify:
   - Key factual claims (what world-facts does it treat as true?)
   - Reasoning steps (how does it proceed logically?)
   - Underlying assumptions (what unstated premises about the world does it rely on?)
   - Intermediate conclusions (what sub-conclusions does it reach?)
   For ARGUMENT B, do the same analysis.
   Then, for each dimension above, explicitly note whether the arguments differ.
2. Distill any differences into cruxes -- questions about objective facts of the world.
   - A crux must be about an objective state of the world -- something that could in principle be resolved by examining reality, not by reading either argument.
   - Do NOT frame cruxes as beliefs, assumptions, or claims of either argument.
     Bad:  "Does the analyst's model assume inflation is sticky?"
     Good: "Is inflation sticky in the relevant period?"
   - Avoid using the conclusion itself as a crux.
   - Write each crux as a neutral question (e.g., "Is X true?").
   - Do NOT mention "Argument A" or "Argument B" in any crux question.
   - You MUST return at least one crux if the arguments rely on different world-facts -- even if they reach the same final answer.
   - Return no cruxes ONLY if the arguments are word-for-word identical or trivial rephrasings with zero substantive difference in reasoning.
   - Identify one crux per genuinely distinct point of disagreement -- don't split a single difference into multiple cruxes, and don't merge separate differences into one.
3. For each crux, produce a per-crux object containing:
   - The crux question (neutral, no mention of "Argument A/B").
   - option_A: Argument A's premise -- a standalone factual claim about the world.
   - option_B: Argument B's premise -- contradicts option_A along the crux dimension.
   - A rationale for each of the four quality criteria explaining why this crux satisfies it.
   Example: If the crux is "Is X true?" then:
       - option_A: "X is true."
       - option_B: "X is not true."
   Do NOT mention "Argument A" or "Argument B" in any premise.
   Each premise must be an objective claim about the world, not a statement about what an argument says or believes.
   The two premises must contradict each other along the dimension the crux asks about.

QUESTION:
{question}

ARGUMENT A:
{argument_A}

ARGUMENT B:
{argument_B}

IMPORTANT: Only identify cruxes grounded in genuine factual disagreements about the world. Do not generate cruxes from stylistic differences, tone, or reasoning methodology. An empty crux list is acceptable if arguments differ only in framing or presentation without disagreeing on any world-fact.

QUALITY REMINDER -- before finalizing each crux, verify it passes all four criteria:
  Binary Contrast: Are option_A and option_B mutually exclusive?
  Intermediate: Is this an upstream sub-question, not a restatement of the original question?
  World Fact: Do both options make objective claims about reality, not about what an argument says?
  Clear: Is it on-topic, specific, well-posed, and easy to understand?
\end{Verbatim}

\paragraph{Metaculus extraction, contrast\_prompts.json id=2 (bottom\_up\_claim\_diffing)}
\begin{Verbatim}[breaklines=true, fontsize=\footnotesize, frame=single, framerule=0.3pt]
You will extract cruxes by systematically diffing the factual claims in two arguments.

QUESTION:
{question}

ARGUMENT A:
{argument_A}

ARGUMENT B:
{argument_B}

=== STEP 1: ATOMIZE ===
Break each argument into numbered atomic claims -- individual factual assertions about the world. Include both explicit claims and implicit assumptions. Each claim should be a single, self-contained factual statement.

For Argument A, list: A1, A2, A3, ...
For Argument B, list: B1, B2, B3, ...

=== STEP 2: DIFF ===
Compare the two claim lists. Classify every pair as one of:
- MATCH: Both arguments assert the same fact.
- CONTRADICT: The arguments assert opposing facts about the same dimension of reality.
- UNIQUE: A claim appears in one argument but has no counterpart in the other.

Focus on CONTRADICT pairs -- these are your crux candidates.

=== STEP 3: PROMOTE TO CRUXES ===
For each CONTRADICT pair (or cluster of related contradictions), formulate a crux question. Apply these quality standards:

- Binary Contrast: The two options must be mutually exclusive -- choosing one rules out the other. Both must be equally specific.
- Intermediate: The crux must be an upstream sub-question, not a restatement of the original forecasting question or its conclusion.
- World Fact: Both options must be objective claims about reality. Do NOT reference what an argument says or believes. No "Argument A thinks X" -- just "X is true" vs "X is not true."
- Clear: The crux must be on-topic, specific, well-posed, and easy to understand. No vague or compound questions.

For each crux, produce:
- question: A neutral question about the world (no mention of Argument A/B).
- option_A: The factual claim from the A-side of the contradiction (standalone).
- option_B: The contradictory factual claim from the B-side (standalone).
- binary_contrast_rationale: Why the two options are mutually exclusive.
- intermediate_rationale: Why this is an upstream sub-question, not a restatement.
- world_fact_rationale: Why both options are objective claims about reality.
- clear_rationale: Why this is specific, on-topic, and well-posed.

Merge related contradictions into a single crux if they address the same underlying factual disagreement. Do not split a single disagreement into multiple cruxes.

Return an empty crux list ONLY if no CONTRADICT pairs exist (the arguments agree on all world-facts). If contradictions exist, you MUST produce at least one crux.
\end{Verbatim}

\subsubsection{Single-forecast extraction prompt}

The single-rollout cruxes (\cref{app:single}) come from this prompt, which reads one forecast and carries the rubric of the March Madness extraction prompt above; \texttt{\{question\}} is the question and \texttt{\{argument\}} one rendered forecast.

\begin{Verbatim}[breaklines=true, fontsize=\footnotesize, frame=single, framerule=0.3pt]
QUALITY RUBRIC -- every crux you produce must satisfy all four criteria:

Binary Contrast: A well-formed crux question poses a clear this-or-that choice. The two options (option_A and option_B) must be mutually exclusive -- choosing one rules out the other.

Intermediate: A good crux addresses an upstream factual sub-question whose answer informs the final forecast. It must NOT be a restatement of the original question or the final answer options. It should identify a specific factual premise that the argument relies on, upstream of its conclusion.

World Fact: Both answer options must make objective claims about the state of the world. They must NOT reference a response, forecast, or forecaster (e.g., "The argument believes X"). Each option should be a standalone factual claim.

Clear: The crux must be on-topic, free of internal contradictions, specific enough to evaluate, and easy to understand.

TASK:
You are given ONE argument: a single forecast for the question below, with its reasoning. Identify the cruxes of this argument: the factual premises about the world that its forecast turns on, so that if a premise were false, the forecast would change.
1. Systematically analyze the argument, identifying:
   - Key factual claims (what world-facts does it treat as true?)
   - Reasoning steps (how does it proceed logically?)
   - Underlying assumptions (what unstated premises about the world does it rely on?)
   - Intermediate conclusions (what sub-conclusions does it reach?)
2. Distill the premises its forecast depends on into cruxes -- questions about objective facts of the world.
   - A crux must be about an objective state of the world -- something that could in principle be resolved by examining reality, not by reading the argument.
   - Do NOT frame cruxes as beliefs, assumptions, or claims of the argument.
     Bad:  "Does the analyst's model assume inflation is sticky?"
     Good: "Is inflation sticky in the relevant period?"
   - Avoid using the conclusion itself as a crux.
   - Write each crux as a neutral question (e.g., "Is X true?").
   - Do NOT mention the argument, the forecast, or the forecaster in any crux question.
   - Identify one crux per genuinely distinct premise -- don't split a single premise into multiple cruxes, and don't merge separate premises into one.
   - Return at most 30 cruxes, most important first: the premises whose reversal would move this forecast the most.
3. For each crux, produce an object containing:
   - The crux question (neutral, no mention of the argument or the forecast).
   - option_A: the premise this argument relies on -- a standalone factual claim about the world.
   - option_B: the contrary premise -- contradicts option_A along the crux dimension, so that if it were true the forecast would change.
   Example: If the crux is "Is X true?" and the argument relies on X, then:
       - option_A: "X is true."
       - option_B: "X is not true."
   Do NOT mention the argument or the forecast in any premise.
   Each premise must be an objective claim about the world, not a statement about what an argument says or believes.
   The two premises must contradict each other along the dimension the crux asks about.

QUESTION:
{question}

ARGUMENT:
{argument}

IMPORTANT: Only identify cruxes grounded in factual premises about the world that this forecast depends on. Do not generate cruxes from stylistic features, tone, or reasoning methodology. Return no cruxes ONLY if the argument relies on no factual premise at all.

QUALITY REMINDER -- before finalizing each crux, verify it passes all four criteria:
  Binary Contrast: Are option_A and option_B mutually exclusive?
  Intermediate: Is this an upstream sub-question, not a restatement of the original question?
  World Fact: Do both options make objective claims about reality, not about what an argument says?
  Clear: Is it on-topic, specific, well-posed, and easy to understand?
\end{Verbatim}

\subsubsection{Scouting-report persona prompts}

The three persona prompts are templates in which \texttt{\%s} is replaced by the team name.
Each call goes to the OpenAI Responses API with model \texttt{gpt-5.4-mini}, the \texttt{web\_search\_preview} tool enabled, and the formatted prompt as \texttt{input}, one call per team per persona.

\paragraph{Analyst persona (Traditional Basketball Analyst)}
\begin{Verbatim}[breaklines=true, fontsize=\footnotesize, frame=single, framerule=0.3pt]
You are a Traditional Basketball Analyst providing a detailed scouting report on %s
for the 2026 NCAA Tournament. Search the web for current 2026 season data.

Cover ALL of the following with specific examples -- no generalities:
- Team record and conference record (exact wins/losses)
- Offensive scheme: specific plays, sets, and tendencies with named examples
- Defensive scheme and philosophy with specific tendencies
- Coaching philosophy and documented in-game adjustment history (cite specific games)
- Clutch and late-game execution: cite specific games with scores and dates
- Key players: role, scoring average, recent form, standout moments (named stats)
- Notable wins AND losses with context (opponent, margin, what it revealed)
- Head coach's NCAA tournament history (rounds reached, years)

Every claim must be backed by a specific figure, named player, or named game.
Minimum 600 words.

Return ONLY valid JSON: {"text": "<full report prose>"}
\end{Verbatim}

\paragraph{Forecaster persona (Quantitative Forecaster)}
\begin{Verbatim}[breaklines=true, fontsize=\footnotesize, frame=single, framerule=0.3pt]
You are a Quantitative Forecaster providing a detailed statistical profile of %s
for the 2026 NCAA Tournament. Search the web broadly for current 2026 season data.

Search for and report whatever you can actually find -- rankings, efficiency metrics, shooting
splits, pace, schedule strength, recent form. Use exact values where available; when a specific
figure isn't accessible, synthesize from what you did find (e.g. 'ranked in the top 10 on KenPom'
is fine if you can't get the exact number). Do not refuse or hedge -- report what the data shows.

Cover as many of the following as your search yields:
- KenPom rank, BPI rank, NET rank
- Adjusted offensive and defensive efficiency (points per 100 possessions)
- Pace (possessions per 40 minutes)
- eFG%, TO%, OR%, FT rate
- 3-point attempt rate
- Strength of schedule
- Efficiency trend over the last 10 games
- Where the numbers diverge from conventional wisdom or the eye test

Synthesize the data into a coherent analytical narrative. Be specific and substantive.
Minimum 600 words.

Return ONLY valid JSON: {"text": "<full report prose>"}
\end{Verbatim}

\paragraph{Matchup analyst persona (Matchup Analyst)}
\begin{Verbatim}[breaklines=true, fontsize=\footnotesize, frame=single, framerule=0.3pt]
You are a Matchup Analyst providing a detailed tactical breakdown of %s
for the 2026 NCAA Tournament. Search the web for current 2026 season data.

Cover ALL of the following with specific examples -- no generalities:
- Style of play and which specific tactics disrupt it (cite examples)
- Press and zone vulnerability with named games as evidence
- Preferred pace and how to force the opposite (specific examples)
- Size and athleticism mismatches at each position (named players)
- Isolation vs. team offense tendencies (with percentages)
- Transition offense and defense frequency (with stats)
- Coaching adaptability: does the staff make adjustments mid-series? (cite examples)
- Concrete gameplan: 'To beat this team you must...' (specific, actionable)

Every claim must be backed by a specific stat, named player, or named game.
Minimum 600 words.

Return ONLY valid JSON: {"text": "<full report prose>"}
\end{Verbatim}
\end{multicols}

\clearpage
\section{Examples}\label{app:examples}

\subsection{Example scouting reports}\label{app:reports}

Each team has three scouting reports generated by gpt-5.4-mini with web search, one from each of the three report perspectives (\cref{sec:marchmadness}).
Below are the three reports for Iowa State, the team in the left panel of \cref{fig:composite_venns}.
Reports are shortened to about a seventh of their length and condensed in our own words, keeping the model's figures, and they disagree on some facts, such as Iowa State's record.
In the released data the statistician perspective is named forecaster.

\raggedcolumns
\begin{multicols}{2}
\begingroup\setlength{\leftmargini}{0pt}
\begin{Forecast}[Iowa State, analyst report]
Iowa State enters the 2026 NCAA Tournament at 27--7 overall and 12--6 in Big 12 play. The offense runs through decision-making guards and physical forwards. Joshua Jefferson leads with 167 assists from the frontcourt, Tamin Lipsey has 156 assists at point, and Blake Buchanan shoots 64.5\% from the field. Milan Momcilovic is the spacing piece with 127 made threes at 49.6\% --- the clearest proof the half-court offense is built around drive-kick action into trail/slot threes.

Against Arizona State on March 7, Iowa State trailed 41--37 at halftime, then ripped off a 24--0 second-half run. Iowa State forced 23 turnovers in that game and 23 more in their Big 12 Tournament rematch (91--42), for 46 combined forced turnovers across two meetings. Their counter-blueprint was exposed at Arizona on March 2 (73--57 loss): Arizona won the glass 40--33 and held ISU to 16 paint points vs.\ their 38.0 season average.

Key players: Momcilovic at 17.1 ppg, 49.6\% from three; Jefferson at 16.9 pts/7.6 reb/167 assists; Lipsey at 13.3 pts/156 assists/68 steals; Buchanan at 8.4 pts/64.5\% FG.
\end{Forecast}

\begin{Forecast}[Iowa State, statistician report]
Iowa State's predictive rankings as of March 10, 2026: No.\ 8 NET, No.\ 12 BPI, No.\ 7 KenPom, No.\ 9 Bart Torvik T-Rank; a separate KenPom-derived snapshot had them No.\ 6 overall with the 21st-ranked offense and 4th-ranked defense. Season record: 27--7, 16--1 at home, 5--5 on the road. Scoring margin: +16.6 (81.8 ppg scored, 65.1 allowed). Shooting: 49.0\% FG, 38.7\% from three; held opponents to 42.6\%/32.1\%. Turnover margin: +5.0 (forced 15.3 per game, committed only 10.3).

Possession-based offensive profile: 56.47\% eFG, 12.91\% turnover rate, 35.32\% offensive rebounding rate, 38.77\% three-point attempt rate. Strength of schedule: No.\ 28 nationally.

A key hidden weakness: Iowa State ranks 340th nationally at 67.2\% from the free-throw line --- a real March vulnerability in coin-flip games. Over the previous 10 games before the tournament, their offensive efficiency had fallen to 121.3 (46th nationally). However, Big 12 tournament results (91--42, 75--53 wins before an 82--80 final loss to Arizona) confirmed the floor-ceiling dynamic remains intact. Bottom line: legitimate second-weekend-or-better team with elite defensive credibility and excellent turnover control. Caution flags: merely good adjusted offense, late-season offensive fade, and poor foul shooting.
\end{Forecast}

\begin{Forecast}[Iowa State, matchup-analyst report]
Iowa State enters the tournament at 26--6 with a 122.3 offensive rating and 96.9 defensive rating. Team stats: 82.1 ppg, 48.9\% FG, 17.4 assists/game, 9.9 turnovers/game, 9.0 steals/game. Style: pressure defense flowing into organized offense. The 17.4 assists vs.\ 9.9 turnovers profile shows a team built on drive-and-kick and frontcourt facilitation, not isolation. Jefferson is the key connector.

What disrupts them: (1) Physical rim pressure from bigger scoring wings --- in BYU's 79--69 win on Feb.\ 21, AJ Dybantsa went for 29/10/9 and BYU outscored ISU 22--6 in the paint in the first half. (2) Dominant rebounding/paint volume --- BYU survived 29 turnovers and still won by going +28 on the glass (52--24). Zone vulnerability is plausible if it takes away middle catches and runs Momcilovic off rhythm.

Five-part game plan to beat Iowa State: (1) Attack the rim with a big initiator before pressure loads. (2) Do not feed their steal game with lazy slot passes. (3) Front Jefferson's elbow catches to make him a scorer rather than a hub. (4) Top-lock Momcilovic and force him off rhythm. (5) Win the rebounding/finishing battle. The wrong plan: trying to out-pressure them.
\end{Forecast}
\endgroup

\end{multicols}

\subsection{One Metaculus rollout}\label{app:rollout}

Below is the Metaculus rollout for the Frontier Group Holdings bankruptcy question of 2026-03-01, made by gpt-5-nano with OpenAI web search.
The seeded single-rollout rule selects this rollout.

\raggedcolumns
\begin{multicols}{2}

\paragraph{Question and metadata}
\begin{Verbatim}[breaklines=true, fontsize=\footnotesize, frame=single, framerule=0.3pt]
Will Frontier Group Holdings, Inc. file for bankruptcy protection before May 1, 2026?

Rollout: one of 30 stored rollouts
Forecast-challenges brief: null
\end{Verbatim}

\paragraph{Forecasts brief}
\begin{Verbatim}[breaklines=true, fontsize=\footnotesize, frame=single, framerule=0.3pt]
Anchor probability: 0.45826666666666666
Confidence: high

Reference class:
Publicly traded U.S. airline companies with high leverage and distress signals (airlines with Altman Z-score in the distress zone, i.e., Z <= 1.8), exemplified by Frontier Group Holdings (ULCC) with an Altman Z-score of 0.38 as of Feb 13, 2026.

Sources:
[12 sources: 6 on the Z-score/ULCC bankruptcy-rate anchor, 4 on Spirit Airlines' Chapter 11, 1 on Frontier's liquidity, 1 on the Q4 2025 earnings webcast; list omitted, see the data release]
\end{Verbatim}

\paragraph{Causal brief}
\begin{Verbatim}[breaklines=true, fontsize=\footnotesize, frame=single, framerule=0.3pt]
Confidence: high

Causal factors:
1. Frontier Altman Z-Score distress signal (direction: up, magnitude: strong)
   Frontier's Altman Z-Score of 0.38 as of February 13, 2026 places it in the distress zone (Z <= 1.8), implying meaningful bankruptcy risk within the next two years.
2. Frontier liquidity cushion and debt position (direction: down, magnitude: moderate)
   Frontier reported roughly $889 million of available liquidity against about $507 million of debt with a cash runway to meet needs for the next twelve months, reducing near-term bankruptcy risk.
3. Peer airline distress and industry dynamics (Spirit Airlines) (direction: up, magnitude: strong)
   Spirit Airlines filed for Chapter 11 in 2024 and ongoing restructuring, coupled with Frontier's takeover efforts, signals industry-wide stress that elevates Frontier's bankruptcy risk relative to peers.
4. Public absence of a Frontier bankruptcy filing by the target date (direction: down, magnitude: moderate)
   There was no public Frontier bankruptcy filing by May 1, 2026 according to corporate disclosures, which tempers risk relative to the anchor.
[entries 5 to 11 omitted; they restate these four factors; see the data release]
\end{Verbatim}

\paragraph{Persona estimate: outside-view}
\begin{Verbatim}[breaklines=true, fontsize=\footnotesize, frame=single, framerule=0.3pt]
Probability: 0.08 (range 0.04 to 0.12, confidence: low)

Narrative:
Starting from a high-distress reference class, Frontier's strong liquidity runway and absence of a bankruptcy filing by 5/1/2026 modestly offset peer-driven pressures. The net Bayes-updating yields a low posterior probability of bankruptcy, acknowledging notable uncertainty in airline industry dynamics.

Summary:
- Posterior probability of Frontier filing for bankruptcy before May 1, 2026 is calibrated at 8% based on Bayesian synthesis of distress signals and liquidity cues.
- Anchor draws on ULCC-style distress in the airline sector (distress indicators and ~45% bankruptcy probability), but Frontier-specific factors dampen the risk.
- Industry peer distress (Spirit Airlines) increases risk signals, yet Frontier's liquidity cushion and absence of a filing as of May 1, 2026 temper the likelihood.
- Overall, risk is elevated relative to general norms but remains modest, with substantial uncertainty about near-term outcomes.

Justification:
Anchor probability 0.35: [basis omitted; see the data release]
Factor updates:
1. Frontier Altman Z-Score distress signal (direction: up, magnitude: strong)
2. Frontier liquidity cushion and debt position (direction: down, magnitude: moderate)
3. Peer airline distress and industry dynamics (Spirit Airlines) (direction: up, magnitude: strong)
4. Public absence of a Frontier bankruptcy filing by the target date (direction: down, magnitude: moderate)
[factor rationales omitted; see the data release]
Synthesis rationale:
The final 8% posterior emerges from balancing strong distress signals (up) with mitigating factors (down). Although peer distress raises risk, Frontier's liquidity and lack of filing provide a downward bias; correlated factors and limited projection horizon justify a cautious, low-probability conclusion with substantial uncertainty.
\end{Verbatim}

\paragraph{Persona estimate: inside-view}
\begin{Verbatim}[breaklines=true, fontsize=\footnotesize, frame=single, framerule=0.3pt]
Probability: 0.08 (range 0.04 to 0.12, confidence: low)

Narrative:
Bayesian updates reduce the anchor from peer distress to an 8% posterior for Frontier by May 1, 2026, mostly due to Frontier's liquidity cushion and lack of a filing by the date, though Spirit-driven industry distress keeps the risk non-zero. The estimate remains sensitive to external shocks in the airline sector and Frontier's strategic decisions.

Summary:
- Posterior probability of Frontier Group Holdings filing for bankruptcy before May 1, 2026 is 8% (0.08).
- Anchor based on ULCC distress reference class with a 24-month bankruptcy odds around 45.7%, used as the starting point for updates.
- Updates from liquidity cushion and absence of filing dampen risk, but peer distress (Spirit Airlines) preserves non-trivial risk, yielding a low but non-zero probability.

[justification omitted; see the data release]
\end{Verbatim}

\paragraph{Persona estimate: contrarian}
\begin{Verbatim}[breaklines=true, fontsize=\footnotesize, frame=single, framerule=0.3pt]
Probability: 0.46 (range 0.40 to 0.54, confidence: medium)

Narrative:
Anchored near the ULCC-based distress benchmark (approximately 45.7% over 24 months). Distress signals from Frontier and its peers argue for a higher risk than average, but liquidity strength and the lack of a Frontier filing by May 1, 2026 reduce the risk, yielding a mid-range probability with substantial uncertainty.

Summary:
- Frontier shows distressed signals (Z-score 0.38) and peer-bankruptcy dynamics (Spirit), elevating bankruptcy risk above baseline.
- Near-term liquidity cushion and no filing by May 1, 2026 offset the risk somewhat, dampening probability.
- Overall, probability remains elevated relative to historical norms but not certain; calibrated around the anchor with uncertainty bands.

[justification omitted; see the data release]
\end{Verbatim}

\paragraph{Final combination}
\begin{Verbatim}[breaklines=true, fontsize=\footnotesize, frame=single, framerule=0.3pt]
Ensemble aggregate
Aggregate probability: 0.1568732819005304
Range: [0.08, 0.46]

Bayes update
Prior: 0.46
Posterior: 0.07849829351535834
Raw posterior (before capping): 0.07848946925772159
Capped: true

Factors (name, P(evidence|yes), P(evidence|no), Bayes factor):
1. Industry distress pressure from Spirit's Chapter 11 -- 0.7 / 0.5 -- bf 1.4
2. Anchor risk elevated by peer distress -- 0.77 / 0.45 -- bf 1.711111111111111
3. Frontier liquidity cushion and debt position -- 0.3 / 0.85 -- bf 0.35294117647058826
4. Public absence of Frontier bankruptcy filing by May 1, 2026 -- 0.1 / 0.92 -- bf 0.10869565217391304
5. Frontier Altman Z-Score distress signal -- 0.84 / 0.16 -- bf 5.25
6. Frontier near-term liquidity runway and cash coverage -- 0.28 / 0.8 -- bf 0.35000000000000003
7. Peer airline distress and industry dynamics -- 0.8 / 0.4 -- bf 2.0
8. Distress signals in Frontier's reference class -- 0.75 / 0.4 -- bf 1.875
9. Company disclosures indicating no Frontier bankruptcy filing as of May 1, 2026 -- 0.15 / 0.95 -- bf 0.15789473684210525

Final probability: 0.1

Summary:
- The ensemble aggregate is 16%, but objections highlight near-term resilience from Frontier's liquidity cushion and absence of a bankruptcy filing as of May 1, 2026, suggesting the risk is overstated.
- Distress signals remain (low Altman Z-score in the distress zone and peer airline distress, e.g., Spirit), so the risk is non-negligible but not imminent.
- Threshold analysis indicates a possible but not highly likely filing before the target date; no evidence of a breach by that date dampens the forecast.
- A modest downward recalibration to about 10% reflects the balance of these factors while preserving some due to industry dynamics.

Justification:
Ensemble aggregate: 0.16
Critique assessment: The objections are valid: they expose legitimate near-term cushions and absence of filing, but distress signals and industry dynamics prevent a collapse to near-zero risk.
Adjustment: Adjusted downward from 0.16 to 0.10 to reflect the strong liquidity cushion and absence of filing by the target date, while recognizing non-zero risk from distress signals and strategic options.
\end{Verbatim}

\end{multicols}

\subsection{The figure 1 demonstration}\label{app:fig1demo}

\cref{fig:overview} illustrates the pipeline on a question outside Metaculus and March Madness.
Claude Opus 5.5 produced 30 forecasts of whether top-model API prices halve by the end of 2027, giving 435 pairs, from which the pipeline extracted 86 unique cruxes.
Each forecast ran its own web searches, so the run depends on what the web held at the time, and it is not part of the reproducibility artifacts.
The figure shows four of those cruxes and three of those pairs, chosen to read at a glance rather than sampled at random, and their wordings are shortened from the extracted text.
No result in the paper depends on this run.

\section{Related Work in Detail}\label{app:related_detail}

\subsection{Extended related work}\label{app:related}

\paragraph{Language model introspection and self-knowledge.} Several papers ask whether a language model can report accurately on its own internal states.
\citet{kadavath2022language} show that language models can sometimes identify which questions they are likely to answer correctly.
\citet{lin2022teaching} fine-tune a model to state its confidence in words, and \citet{kapoor2024taught} show that prompting alone does not make such statements calibrated.
A separate line tests whether chain-of-thought reasoning~\citep{wei2022chain} is faithful.
\citet{turpin2024language} show that stated reasoning can be unfaithful to the computation that produced the answer.
\citet{chen2025reasoningmodelsdontsay} find that reasoning models, which write out their thinking before they answer, often leave out what influenced the answer.
\citet{chua2025deepseekr1reasoningmodels} find that reasoning models leave it out less often than models that answer directly.
The multiple-rollout cruxes here are read off written rationales, so they are the cruxes of the reasoning a person reads, whether or not that reasoning is faithful to the computation behind the forecast.
\citet{binder2025looking} show that a model fine-tuned on labels of its own behavior can learn to predict its own tendencies.
\citet{betley2025tell} show that models fine-tuned to exhibit a single behavior can describe that behavior when asked, and \citet{laine2024sad} benchmark what a model knows about itself and its situation.
\citet{bogdan2025thoughtanchorsllmreasoning} identify which individual reasoning steps in a chain of thought have downstream effects on the final answer.
Like the comparison of rollouts here, that method finds after the fact what an answer depends on, at the level of steps inside one chain of thought rather than factual questions across rollouts.
This paper asks instead whether a model, without fine-tuning, can predict the factual questions on which many rollouts of one request will take different sides.\footnote{%
This paper calls a set of cruxes named in advance a prediction rather than introspection, since nothing in the test requires the causal access to internal states that the stricter definition of \citet{lindsey2026emergent} demands.
\citet{comsa2025does} give a looser definition, under which a model that infers its own state from text it has generated counts as introspecting.
The a priori cruxes are written without the predictor seeing any rollout to infer from, so this paper calls them a prediction under either definition.}

\paragraph{Claim decomposition.} Key Point Analysis~\citep{barhaim2020kpa} extracts key points from argument collections.
FActScore~\citep{min2023factscore} and SAFE~\citep{wei2024safe} break a long text into single-fact claims and check each one.
Comparing two rollouts to find cruxes, which this paper offers as a method for extracting cruxes, also breaks text into short factual claims.
It differs from claim decomposition in reading two rollouts at once and returning only the claims on which the two take different sides.

\paragraph{Language model aggregation and consistency.} Self-consistency~\citep{wang2023selfconsistency} samples multiple reasoning paths and takes a majority vote, exploiting diversity across outputs to improve accuracy.
Multi-agent debate~\citep{du2024improving} has models critique each other's outputs to converge on better answers.
This paper instead breaks the disagreement among outputs into cruxes and asks whether a model could have predicted them.

\paragraph{Forecasting with language models.} \citet{halawi2024approaching} build a system in which a language model retrieves news and forecasts questions from forecasting platforms nearly as well as the aggregate of human forecasters.
\citet{murphy2026agentic} build a forecasting agent that updates a written belief, a probability with a summary of the evidence, at each step of a tool-use loop, and aggregates several independent runs into one forecast.
Those runs are combined rather than compared, and the result here suggests that contrasting them would surface the cruxes on which they differ.
The Metaculus rollouts here come from a system of the same kind (\cref{sec:metaculus}), and this paper studies the cruxes on which its rollouts take different sides.

\paragraph{Benchmarking cruxes.} \citet{dai2026cruxbenchbenchmarkinformationdiscovery} introduce CruxBench, a benchmark of information discovery.
For each of 293 resolved prediction-market questions they give eight models the same news articles, ask each for one subquestion that resolves inside a 30-day window, and score that subquestion by the value of information its resolution carries, estimated from the target market's price move around the time the crux resolved.
In the terms used here, every crux in that benchmark is a priori, because no part of it reads cruxes off a model's own forecasts.
CruxBench therefore ranks models by how informative a named crux is, against a market's beliefs and only after the event resolves.
The test here asks whether a model's named cruxes are the ones its own forecasts disagree on, which needs no event to resolve and no market.
Their hindsight ceiling, the best moment in each window chosen after the fact, is 0.129 bits, so the headroom they report is nearly an order of magnitude.

\end{document}